\documentclass[conference]{IEEEtran}

\usepackage{amssymb}
\usepackage{times}
\usepackage{graphicx}
\usepackage{subfig}
\usepackage[cmex10]{amsmath}

\usepackage{multirow}
\usepackage{booktabs}

\usepackage{amsthm}
\newtheorem{theorem}{Theorem}
\newtheorem{proposition}{Proposition}
\newtheorem{definition}{Definition}

\begin{document}




\title{Local Probabilistic Model for Bayesian Classification: a Generalized Local Classification Model}


%
\author{\IEEEauthorblockN{Chengsheng Mao$^{a, b, *}$, Lijuan Lu$^c$, Bin Hu$^b$}
\IEEEauthorblockA{$^a$ Department of Preventive Medicine, Feinberg School of Medicine, Northwestern University, Chicago, IL, USA  \\
$^b$ School of Information Science and Engineering, Lanzhou University, Lanzhou, China  \\
$^c$ Shenzhen Sun \& Lynn Circuits Co., Ltd., Shenzhen, China  \\
$^\ast$Corresponding to Email: chengsheng.mao@northwestern.edu}
}

\maketitle

\begin{abstract}
In Bayesian classification, it is important to establish a probabilistic model for each class for likelihood estimation. Most of the previous methods modeled the probability distribution in the whole sample space. However, real-world problems are usually too complex to model in the whole sample space; some fundamental assumptions are required to simplify the global model, for example, the class conditional independence assumption for naive Bayesian classification. In this paper, with the insight that the distribution in a local sample space should be simpler than that in the whole sample space, a local probabilistic model established for a local region is expected much simpler and can relax the fundamental assumptions that may not be true in the whole sample space. Based on these advantages we propose establishing local probabilistic models for Bayesian classification. In addition, a Bayesian classifier adopting a local probabilistic model can even be viewed as a generalized local classification model; by tuning the size of the local region and the corresponding local model assumption, a fitting model can be established for a particular classification problem. The experimental results on several real-world datasets demonstrate the effectiveness of local probabilistic models for Bayesian classification.
\end{abstract}




\section{Introduction}

Bayesian Classification is a naturally probabilistic method that performs classification tasks based on the predicted class membership probabilities, i.e. the probability that a given sample belongs to each class \cite{duda2012pattern}. With the output membership probabilities, a Bayesian classifier provides a degree of confidence for the classification decision, which is more meaningful than a simple assertion of class label. The class membership probabilities in Bayesian classification are estimated based on Bayes' theorem. By Bayes' theorem, the estimation of a class membership probability is transformed to the estimation of the prior probability and the corresponding conditional probability.

In Bayesian classification, the most important step is to estimate the conditional probability for each class; in multi-variable cases, it usually relates to the joint probability estimation using samples in a certain class. Research efforts have been made to estimate the multivariate joint Probability Density Function (PDF) for Bayesian classification. The naive Bayesian classifiers (NBC) \cite{domingos1996beyond,hand2001idiot} take the class conditional independence assumption to transform the multivariate joint conditional PDF into the product of several univariate conditional PDF, which extremely reduces the computation of joint PDF. However, the effectiveness of NBC quite relies on the Class Conditional Independence Assumption (CCIA) which is rarely true in real-world applications. The Bayesian belief networks \cite{friedman1997bayesian} were proposed to allow the representation of dependence among the features, but to train an unconstrained belief network is computationally intensive; it was shown that probabilistic inference using Bayesian belief networks is NP-hard \cite{cooper1990computational,dagum1993approximating}. Usually, a weaker conditional independence is assumed to train a belief network with the complexity between Naive Bayes and an unconstrained belief network for a compromise, e.g. Tree Augmented Naive Bayes (TAN) \cite{friedman1997bayesian} and Aggregating One-Dependence Estimators (AODE) \cite{webb2005not}. In addition, the non-naive Bayesian classifier (NNBC) \cite{wang2014non} was proposed to estimate the multivariate joint PDF directly using multivariate Kernel Density Estimation (KDE) with an optimal bandwidth.

Most of the previous methods try to establish a Global Probabilistic Model (GPM) in the whole sample space for the joint probability distribution estimation. However, the real-world problems are usually too complex to model an effective GPM in the whole sample space or the corresponding GPM is too complex for efficient classification; some fundamental assumptions are usually required to simplify the GPM, for example, the CCIA for NBC. In fact, for the classification of a certain sample, the GPM is unnecessary, and a Local Probabilistic Model (LPM) that models the local distribution around the query sample is sufficient for the estimation of probability distribution at the exact query point.

In addition, a Bayesian classifier based on LPM is a generalized model that can be specialized to a number of existing classification methods, e.g., k-nearest neighbors (kNN) and NBC. We propose a unified form for all local classification methods, where different classification methods may have different parameters. By tuning the parameters, we can establish a fitting model for a particular classification problem. To our best knowledge, this is the first report that shows a generalization of local classification methods.

There are several obvious advantages of using an LPM for Bayesian classification, as follows:
\begin{itemize}
  \item An LPM established for a local region is expected much simpler and can relax the fundamental assumptions that may not be true in the whole sample space;
  \item By the idea of local learning \cite{bottou1992local,huang2005local,wu2006local}, a certain sample should be more related to its nearby samples, thus an LPM established from more related samples should be more fit for the probability distribution in the local region around the query sample;
 \item Bayesian classification based on LPM is a generalization of several existing classification models. A selective LPM can be flexible for problems with various complexities through tuning the size of the local region and the corresponding local model assumption.
\end{itemize}

\section{Preliminaries and Related Work}\label{sec:2}
\subsection{Bayesian Decision for Classification}
Given a query sample with known feature description $\mathbf{x}=[x_1,\cdots,x_d]$ and unknown class label $l\in Y$, where $Y=\{y_1,\cdots,y_c\}$ is a finite set of possible class labels. A Bayesian classifier estimates the posterior probability of class $l$, $P(l \vert \mathbf{x})$ and predicts the best class label $\hat{l}$ for $\mathbf{x}$ using the optimal decision rule that minimizes the conditional risk as equation \ref{eq:bayes}. \cite{bishop2006pattern}
\begin{equation}
\begin{aligned}
\hat{l} & = \arg \min_{l\in Y} R(l|\mathbf{x})   \\
   & = \arg \min_{l\in Y} \sum_{i=1}^{c} \lambda(y_i,l)P(y_i \vert \mathbf{x})
\end{aligned}
\label{eq:bayes}
\end{equation}
where $\lambda(y,l)$ is the loss function (LF) that states the costs for assigning a true class label $y$ as $l$. In classification problems, the zero-one Loss Function (0-1 LF) is usually assumed if the true LF is unknown. The 0-1 LF assigns no loss to a correct classification decision and uniform unit loss to an incorrect decision, defined as
\begin{equation}\label{eq:0-1LF}
    \lambda_{0-1}(y,l)=
    \begin{cases}
        0, \quad \quad  y=l;  \\
        1,  \quad \quad  y\neq l
    \end{cases}
\end{equation}

With the 0-1 LF, the predicted class can be simplified as
\begin{equation}
\begin{aligned}
\hat{l} & = \arg \min_{l\in Y} \sum_{i=1}^{c} P(y_i \vert \mathbf{x}) - P(l \vert \mathbf{x})  \\
& = \arg \min_{l\in Y} 1-P(l \vert \mathbf{x}) \\
& = \arg \max_{l\in Y} P(l \vert \mathbf{x})
\end{aligned}
\label{eq:01bayes}
\end{equation}

${P}(l \vert \mathbf{x})$ is the posterior probability of class $l$, given the sample $\mathbf{x}$, and can be calculated by Bayes' theorem as
\begin{equation}
\begin{aligned}
P(l \vert \mathbf{x}) & = \frac{P(l)P(\mathbf{x} \vert l)}{P(\mathbf{x})}  \\
 & = \frac{p(l)P(\mathbf{x} \vert l)}{\sum_{j=1}^{c}{P(y_j)P(\mathbf{x} \vert y_j)}}.
 \end{aligned}
\label{posterioryi}
\end{equation}

As $P(\mathbf{x})$ is constant for all classes, the best class label by Bayesian rule is
\begin{equation} \label{eq:bayesrule}
\hat{l} = \arg \max_{l\in Y} P(l)P(\mathbf{x} \vert l).
\end{equation}

Thus, there are two items should be estimated from the training set. Let the training set $T$ be given as $T=\{[\mathbf{x}_1,l_1], \cdots, [\mathbf{x}_n,l_n]\}$; each sample $\mathbf{x}_i$ has $d$ descriptors, i.e. $\mathbf{x}_i=[x_{i,1},\cdots,x_{i,d}]$. $P(l)$ is usually estimated by the corresponding frequency as
\begin{equation}
\hat P(l) = \frac{\sum_{j=1}^{n}{I(l_j=l)}}{n} = \frac{|L|}{n},  \quad l\in Y
\label{prioryi}
\end{equation}
where $L$ is a subset of $T$ such that the class labels of samples in $L$ are all $l$.

$P(\mathbf{x} \vert l)$ is the likelihood of sample $\mathbf{x}$ related with class $l$ and can usually be estimated as a multivariate joint probability distribution of class $l$ at $\mathbf{x}$ using the training samples in $L$. If discrete features exist in the $d$ descriptors, $P(\mathbf{x}|l)$ can also be transformed to conditional joint PDF estimation for continuous variables by
\begin{equation}\label{eq:difeature}
    P(\mathbf{x} \vert l) = P(\mathbf{x}_D,\mathbf{x}_C | l) = P(\mathbf{x}_C | \mathbf{x}_D,l)P(\mathbf{x}_D | l)
\end{equation}
where $\mathbf{x}_C$ and $\mathbf{x}_D$ is the continuous and discrete component in $\mathbf{x}$, respectively. Thus, in this paper, we mainly focus on the multivariate joint probability density estimation for continuous variables for Bayesian classification.

\subsection{Related Work}

Although literature on using local probabilistic models for density estimation can be found at \cite{hjort1996locally,loader1996local,vincent2003locally}; however, previous methods mainly focus on univariate density estimation with large sample sizes and would incur dimensionality curse when applied to classification problems with multi-features. The idea of localization for classification can also be found. The \textit{k}-nearest neighbor (kNN) methods that classify a query sample based on a majority voting of its $k$ nearest neighbors should be the original local method for classification \cite{cover1967nearest,larose2006k}. Research on Support Vector Machines (SVM) constructed in a local region can be found in \cite{zhang2006svm,cheng2007localized,blanzieri2008nearest,ladicky2011locally}. Also, some Bayesian classifiers try to train a series of local models based on a subset of features, including NBtree \cite{kohavi1996scaling} and Lazy Bayesian Rules (LBR) \cite{zheng2000lazy}. Research on  building a naive Bayesian classifier in a neighborhood includes \cite{frank2002locally,xie2002snnb,hu2015bayesian}. However Limited research on Bayesian classification using local probabilistic models for likelihood estimation can be found, several local classification methods can be a specification of LPM-based Bayesian classification, see Section \ref{subsec: specification}.

\section{Local Probabilistic Model}\label{sec:3}
For a probability density estimation problem, a simple parametric model can not always describe the complex distribution in the real world; while a non-parametric model usually requires quite a lot of samples for an effective estimation, especially in high dimensional cases. As can be expected that the probability distribution in a local region should not be so complex and can be estimated through a simple parametric model. Thus, it should be a feasible solution for probability density estimation through an LPM. In this section, we will introduce the concept of local probability distribution and the corresponding local probabilistic model.

\subsection{Local Probability Distribution}
For a random variable\footnote{It can be a single variable for univariate distribution or a multi-variable for multivariate joint distribution and so is it later.}, we use local probability distribution to describe the probability distribution in a subset of the sample space. To facilitate a better understanding, we make the following definition.

\begin{definition}[Local Probability]
Suppose $\delta$ and $R$ are two subsets of the sample space of a random variable $\mathbf{x}$, the local probability of $\delta$ in local region $R$ is defined as the conditional probability of $\mathbf{x} \in \delta$ given that $\mathbf{x} \in R$, denoted as
\begin{equation}
 P_R(\delta)=P(\delta \vert R).
\end{equation}
\end{definition}

For a continuous random variable, the local probability density is also defined.
\begin{definition}[Local Probability Density (LPD)]
In the sample space of a continuous random variable $\mathbf{x}$, given a continuous closed region $R$, for an arbitrary point $\mathbf{x}_0$, the local probability density in $R$ is defined as
\begin{equation}
 f_R(\mathbf{x}_0)=\lim_{\delta(\mathbf{x}_0)\rightarrow \{\mathbf{x}_0\}}\frac{P_R(\delta(\mathbf{x}_0))}{V(\delta(\mathbf{x}_0))}
\end{equation}
where $\delta(\mathbf{x}_0)$ and $V(\delta(\mathbf{x}_0))$ respectively denote the neighborhood of $\mathbf{x}_0$ and its volume\footnote{$V(\cdot)$ denotes the size of a region, volume for 3-dimensional cases, area for 2-dimensional cases and hyper-volume for higher dimensional cases.}.
\end{definition}

Similar to the global probability density to the whole sample space, the LPD describes the relative likelihood for a random variable to take on a certain point given that it is in a certain local region. If the local region is extended to the whole sample space, the local distribution becomes the general global distribution, and LP/LPD should become the conventional probability/probability density. Similar to the conventional probability density, LPD also satisfies the nonnegativity and unitarity in the corresponding local region.
\begin{proposition}
If $f_R(\mathbf{x})$ denotes the local probability density of a continuous random variable $\mathbf{x}$ in a continuous closed region $R$, then $f_R(\mathbf{x})$ has following two properties.\\
(I) Nonnegativity:
\begin{equation}
\ f_R(\mathbf{x})
\begin{cases}
\geq 0, \quad  for \quad \mathbf{x} \in R;  \\
= 0, \quad  for \quad \mathbf{x} \notin R
\end{cases}
\end{equation}
(II)  Unitarity :
\begin{equation} \label{unitarity}
\int_{R}{f_R(\mathbf{x})d\mathbf{x}}=1.
\end{equation}
\end{proposition}

The following proposition can describe the relationship between the general global PDF and LPD.
\begin{proposition} \label{pdfest}
In the sample space of a continuous random variable $\mathbf{x}$, if a continuous closed region $R$ has the prior probability $P(R)$, and the local probability density function for region $R$ is $f_R(\mathbf{x})$, then the global probability density of a point $\mathbf{x}_0\in R$ is
\begin{equation}
{f(\mathbf{x}_0)=f_R(\mathbf{x}_0)P(R)} \quad  for \quad \mathbf{x}_0\in R.
\label{globalpdf}
\end{equation}

\end{proposition}

Proposition \ref{pdfest} provides a method to estimate global density from LPD, which is the basic idea of our method. Given an arbitrary point $\mathbf{x}$ in the sample space, we consider a local region $R$ containing $\mathbf{x}$; if the prior probability $P(R)$ is known or can be estimated, the estimation of global density is transformed to the estimation of LPD.

\subsection{Local Model Assumption}
The estimation of LPD is based on the samples falling in the corresponding local region; it is similar to the estimation of PDF in the whole sample space, except that the local distribution is supposed to be much simpler. Thus we can assume a simple probabilistic model in the local region for the LPD estimation. Suppose we totally have $n$ observations among which there are $k$ samples falling in the local region $R$, denoted by $\mathbf{x}_1,\cdots,\mathbf{x}_k$, the following local model assumptions can be taken for the LPD estimation.

\subsubsection{Locally Uniform Assumption}
The Locally Uniform Assumption (LUA) assumes a uniform distribution in the local region. Thus, according to the unitarity of LPD in region $R$ (Formula \ref{unitarity}), the LPD with LUA is estimated as
\begin{equation}
\hat f_R(\mathbf{x}) = \frac{1}{V(R)}, \quad for \quad \mathbf{x} \in R.
\label{LUALPD}
\end{equation}

With the prior probability of $R$ estimated by $\hat P(R)=k/n$, the PDF can be estimated as $\hat{f}(\mathbf{x})=\frac{k}{nV(R)}$, in agreement with the histogram estimator. The histogram estimator can achieve an accurate estimation when $k, n\rightarrow+\infty$ and $V(R)\rightarrow 0$. It can also be interpreted by LUA in a local region; $V(R)\rightarrow 0$ can guarantee an accurate estimation of $f_R(\mathbf{x})$; and $\hat P(R)=k/n$ is more accurate with a larger number of samples.

However, if the sample size is not large enough, a large $k$ cannot coexist with a small local region in practice. To achieve a more accurate estimate for $P(R)$, the region $R$ should be somewhat large and thus cannot be always regarded uniform, and reasonably, a more complex distribution should be assumed.

\subsubsection{Locally Parametric Assumption}
If the local region is not small enough to take the simple LUA, we may assume a somewhat complex parametric model to model the local distribution, termed Local Parametric Assumption (LPA). The LPA assumes a parametric model $f(\mathbf{x};\theta)$ can model the distribution in the local region $R$ with an unknown parameter vector $\theta$. An estimate of $\theta$ can be obtained from the samples in $R$. Note that the LPD should obey the unitarity in $R$ as Formula \ref{unitarity}. Thus, the LPD in $R$ should be normalized as
\begin{equation}
\hat f_R(\mathbf{x})=\frac{f(\mathbf{x};\hat\theta)}{\int_{R}{f(\mathbf{x};\hat\theta)d\mathbf{x}}}, \quad for \quad \mathbf{x} \in R.
\label{lpdest}
\end{equation}

Specifically, we can assume samples in region $R$ follow a Gaussian distribution $g(\mathbf{x};\mu,\Sigma)$ as
\begin{equation}\label{eq:gassianfun}
g(\mathbf{x};\mu,\Sigma)=\frac{1}{\sqrt{(2\pi)^d|\Sigma|}} \exp\{-\frac{(\mathbf{x}-\mu)^T\Sigma^{-1}(\mathbf{x}-\mu)}{2}\}
\end{equation}
and the unknown parameters mean $\mu$ and/or covariance matrix $\Sigma$ can be estimated as
\begin{equation}\label{eq:mean}
    \hat{\mu}=\frac{1}{k}\sum_{i=1}^k{\mathbf{x}_i}
\end{equation}
\begin{equation}\label{eq:sigma}
    \hat{\Sigma}=\frac{1}{k}\sum_{i=1}^k{(\mathbf{x}_i-\hat{\mu})(\mathbf{x}_i-\hat{\mu})^T}
\end{equation}
We term this assumption locally Gaussian assumption (LGA).

\subsubsection{Locally Complex Assumption}
Although the local distribution can be supposed simpler than the global distribution, however, if the original global distribution is too complex and the local region may not be small enough to take the LUA and LPA to model the local distribution, we may take the locally complex assumption (LCA) and employ the non-parametric method for LPD estimation. Similarly, the KDE can also be employed for the LPD estimation with an assigned bandwidth vector $\mathbf{h}$. KDE estimates a global distribution from the samples in local region $R$ as
\begin{equation}
\hat{f}(\mathbf{x})=\frac{1}{k \cdot prod(\mathbf{h})}\sum_{i=1}^{k}{K((\mathbf{x}-\mathbf{x}_i)./\mathbf{h})}
\label{kde}
\end{equation}
where the operator $./$ denotes the right division between the corresponding elements in two equal-sized matrices or vectors; $prod(\cdot)$ returns the product of the all elements in a vector; and $K(\cdot)$ is the kernel function of which the Gaussian kernel is usually used as
\begin{equation}
K(\mathbf{x})=\frac{1}{(2\pi)^{d/2}}\exp\{-\frac{\mathbf{x} \cdot \mathbf{x}^\top}{2}\}.
\label{kernel}
\end{equation}
where $d$ is the dimensionality of $\mathbf{x}$.

And then the global distribution is normalized to the region $R$ for LPD estimation as
\begin{equation}
\hat f_R(\mathbf{x})=\frac{\hat{f}(\mathbf{x})}{\int_{R}{\hat{f}(\mathbf{x})d\mathbf{x}}}, \quad for \quad \mathbf{x} \in R.
\label{KDELPD}
\end{equation}

\subsection{Analysis}
In terms of the complexities of a model, it is usually described by the number of effective parameters \cite{trevor2009elements}. Give a certain local region, LUA has no parameters to estimate, it is the simplest LPM. LPA assumes a single parametric model with its intrinsic parameter to estimate.  While LCA virtually assumes a mixture model with $k$ single parametric models, thus its complexity is the sum of $k$ single models. Therefore, in terms of the complexity, LUA $<$ LPA $<$ LCA.

The local distribution should be much simpler than the global distribution and can be modeled easily. The simplicity can be reflected in two aspects: (1) the relationship between features in a local region can be simpler than that in a larger region containing the local region; and (2) for a single feature, the local model is not more complex than the global model.

The relationship between features can be described by the following theorem.
\begin{theorem} \label{independence}
If a set of random variables are independent in a region $\Omega$, then these variables are independent in an arbitrary region $\omega \subseteq \Omega$ where the variables in the boundary of $\omega$ are independent.
\end{theorem}

Theorem \ref{independence} indicates that feature independence in a local region is a weaker constraint than that in the whole sample space. The independence assumption that may not be true in the whole sample space may be applicable in a local region. Thus, due to the more likely local independence, the joint PDF estimation can be simplified by naive Bayesian in a local region. Actually, the independence assumption between features reduces the number of parameters of a multivariate probabilistic model by assuming the correlation between features as zero.

\begin{figure}[!t]
\centering
\subfloat[Uniform mixture]{\includegraphics[width=0.5\columnwidth]{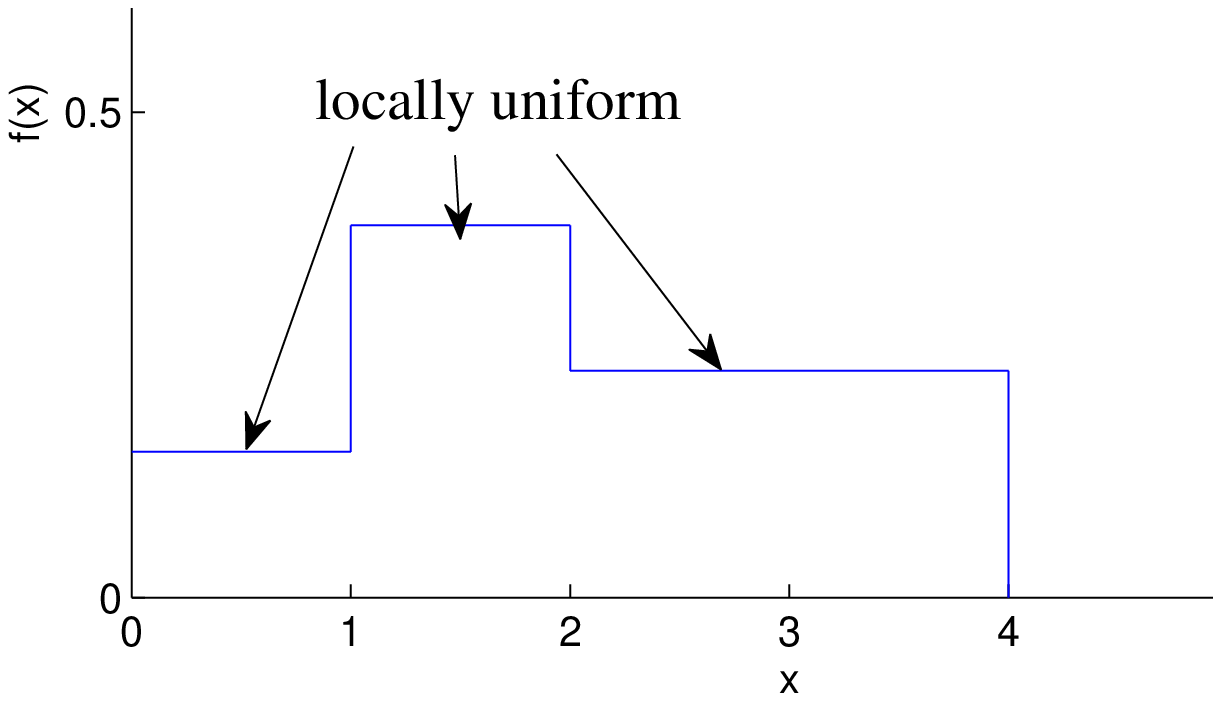}
\label{mixuniform}}
\subfloat[Gaussian mixture]{\includegraphics[width=0.5\columnwidth]{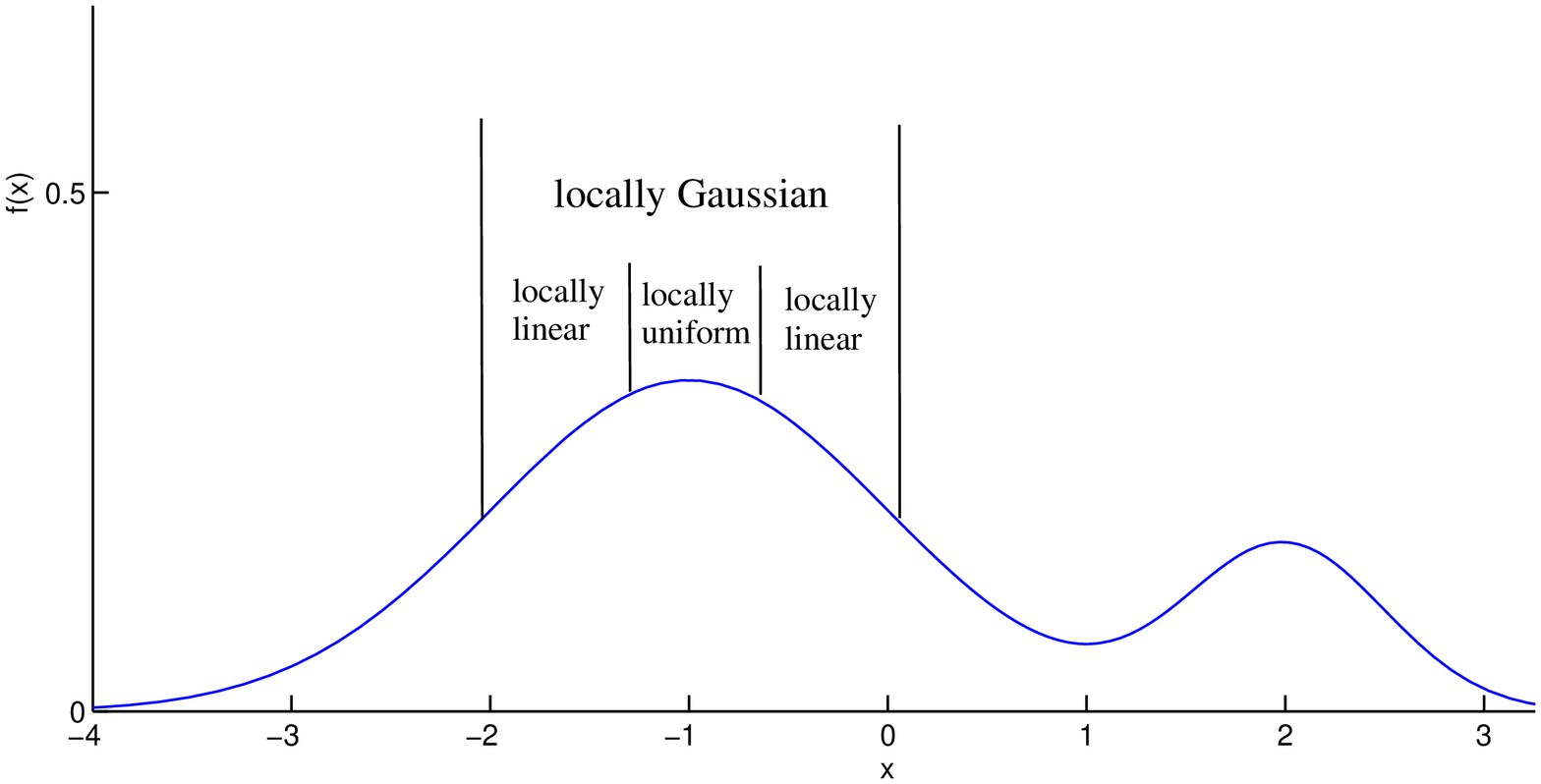}
\label{mixgaussian}}
\caption{The local distribution is simpler than the original global distribution.}
\label{example}
\end{figure}

Additionally, for a single variable, the local model cannot be more complex, and usually simpler, than the original global model. For example, in Figure \ref{mixuniform} the original global distribution is a uniform mixture distribution model $f(x)=0.3U(0,2)+0.7U(1,4)$, where there are 5 parameters; it is a piecewise uniform distribution in the corresponding local region, simpler than the mixture distribution. In Figure \ref{mixgaussian} the original global distribution is a Gaussian mixture distribution $f(x)=0.8N(-1,1)+0.2N(2,0.5)$, where there are also 5 parameters; the local distribution can be approximately uniform (no parameter), linear (one parameter) and Gaussian (two parameters), all simpler than the mixture Gaussian distribution.

From Equation \ref{globalpdf}, the effectiveness of the PDF estimation depends on the estimation of LPD and the prior probability of the corresponding local region, so it is very important to selected the local region and the local model assumption. If the local region $R$ is too small, the prior estimation of $\hat P(R)=k/n$ should have a larger relative error; the extreme case would happen if there are no samples in the local region, where the prior estimate is 0, and ineffective for PDF estimation. Conversely, if we choose a large local region $R$, though the estimation of $P(R)$ is more accurate, the LPM may be too complex for an effective estimate of $f_R(\mathbf{x})$; when the local region is extended to the whole sample space, it becomes the general global PDF estimation.

We provide a criterion for local region selection; while ensuring the accuracy of the estimation of $P(R)$, try to choose the smallest local region where the LPM should be simple. Thus, the local region selection should be related with the sample size $n$, and the selection of the LPM is related with the local region. If $n$ is sufficiently large, the estimation of $\hat P(R)=k/n$ can be effective even in a relative small local region where the local distribution is simple and can be assumed uniform. That is the reason why the histogram estimator is effective when $n \rightarrow +\infty$. However, the sample size $n$ in practice is small, especially in high dimensional cases. In the case of small sample size, to ensure the accuracy of $\hat P(R)=k/n$ will produce a relatively large local region where the LUA does not hold; a more complex local model assumption (LPA or LCA) should be taken.

\section{Classification Rules}\label{sec:4}
Because the global probability density can be effectively estimated through assuming an appropriate LPM, a Bayesian Classifier based on LPM (LPM-BC) can be constructed.

\subsection{Formulation}
Given a query sample $\mathbf{x}$ with unknown class label $l\in Y$, where $Y=\{y_1,\cdots,y_c\}$ is a finite set of possible class labels, a Bayesian classifier estimates posterior probability $P(l \vert \mathbf{x})$ for each class label $l\in Y$,  and predicts the best class label $\hat{l}$ for $\mathbf{x}$ by minimizing the posterior probability based on Bayes' theorem, as
\begin{equation}\label{eq:bayesrule}
\begin{aligned}
\hat{l} &= \arg \max_{l\in Y} {P}(l \vert \mathbf{x})   \\
 & = \arg \max_{l\in Y} {P}(\mathbf{x} \vert l)P(l).
\end{aligned}
\end{equation}

According to Formula \ref{globalpdf}, the probability density at point $\mathbf{x}$, for class $l$ can be extended as
\begin{equation}\label{eq:likelihood}
    P(\mathbf{x}|l)=P(R|l)f_R(\mathbf{x}|l), \quad \mathbf{x} \in R
\end{equation}

For the classification of a certain sample $\mathbf{x}$, a local region $R_l(\mathbf{x})$ for each class $l$ should be determined such that $\mathbf{x}\in R_l(\mathbf{x})$ ($l=y_1,\cdots,y_c$). And then the Bayesian classification rule Equation \ref{eq:bayesrule} can be further transformed to
\begin{equation}\label{eq:localbayesrule}
\begin{aligned}
    \hat{l} & = \arg \max_{l\in Y} P(l)P(R_l(\mathbf{x})|l)f_{R_l(\mathbf{x})}(\mathbf{x}|l)  \\
    & = \arg \max_{l\in Y} P(l,R_l(\mathbf{x}))f_{R_l(\mathbf{x})}(\mathbf{x}|l)
\end{aligned}
\end{equation}

The output posterior probability can also be computed according to the law of total probability as
\begin{equation}
P(l \vert \mathbf{x}) = \frac{P(l,R_l(\mathbf{x}))f_{R_l(\mathbf{x})}(\mathbf{x}|l)}{\sum_{l\in Y}{P(l,R_l(\mathbf{x}))f_{R_l(\mathbf{x})}(\mathbf{x}|l)}}.
\label{eq:localposterior}
\end{equation}

Thus, to classify a certain sample $\mathbf{x}$, we should select a local region $R_l(\mathbf{x})$, then estimate the prior probability $P(l,R_l(\mathbf{x}))$ and assume an LPM in each $R_l(\mathbf{x})$ for the corresponding LPD estimation from the training set.

$P(l,R_l(\mathbf{x}))$ represents the probability that a sample belongs to both class $l$ and the local region $R_l(\mathbf{x})$. If there are $k_l$ out of $n$ samples in class $l$ falling into region $R_l(\mathbf{x})$, it can usually be estimated as
\begin{equation}\label{eq:localprior}
    P(l,R_l(\mathbf{x}))=k_l/n
\end{equation}

Thus,
\begin{equation}\label{grule}
    \hat{l} = \arg \max_{l\in Y} k_lf_{R_l(\mathbf{x})}(\mathbf{x}|l)
\end{equation}
\begin{equation}
P(l \vert \mathbf{x}) = \frac{k_lf_{R_l(\mathbf{x})}(\mathbf{x}|l)}{\sum_{l\in Y}{k_lf_{R_l(\mathbf{x})}(\mathbf{x}|l)}}.
\label{eq:localposterior2}
\end{equation}

The estimation of LPD $f_{R_l(\mathbf{x})}(\mathbf{x}|l)$ depends on the LPM assumption. If an LPM for region $R_l(\mathbf{x})$ and class $l$ is assumed as LUA, LPA or LCA, the corresponding LPD can be estimated from Equation \ref{LUALPD}, \ref{lpdest} and \ref{KDELPD}, respectively.

Due to $\mathbf{x}\in R_l(\mathbf{x})$, $R_l(\mathbf{x})$ is actually a neighborhood of $\mathbf{x}$ and can be selected centered as $\mathbf{x}$ to facilitate the computation related with it. Note that the neighborhood of different samples may overlap and so that the estimated PDF will not be integrated to unity. However, for the classification of a certain sample, we only need the likelihood of that point in each class, the true PDF at every point is not necessary.

\subsection{Specification} \label{subsec: specification}
Formula \ref{grule} can be viewed as a generalized local classification model and can be specialized to a series of different classification rules by selecting various local region $R_l(\mathbf{x})$ and various LPM assumptions.

\textbf{Bayesian rule.} The common Bayesian classification rule is essentially an LPM-BC with the local region $R_l(\mathbf{x})$ extended to the whole sample space, where the LPD becomes the global PDF. Thus in the global case, the LPM-BC rule described by formula \ref{eq:localbayesrule} can be transformed back to the Bayesian rule in Formula \ref{eq:bayesrule}.

\textbf{KNN rule.} If we select an identical neighborhood $R(\mathbf{x})$ for all the classes and take LUA for LPD estimation as $f_{R(\mathbf{x})}(\mathbf{x}|l)=1/V(R(\mathbf{x}))$, constant for all classes, then the LPM-BC rule in Formula \ref{grule} can be transformed to $\hat{l} = \arg \max_{l\in Y} k_l$. That is, it outputs the class that has the most samples in the neighborhood, in agreement with the traditional voting kNN rule.

\textbf{Distance weighted kNN rule (DW-kNN).}
A DW-kNN rule \cite{hechenbichler2004weighted} is essentially a reduced form of LPM-BC with an identical neighborhood $R(\mathbf{x})$ for all the classes and with LCA for LPD estimation. The specialization can be described as
\begin{equation}
\begin{aligned}
\hat{l} & = \arg \max_{l\in Y} k_lf_{R(\mathbf{x})}(\mathbf{x}|l)  \\
& = \arg \max_{l\in Y} k_l \cdot \frac{1}{k_l \cdot prod(\mathbf{h}_l)}\sum_{i=1}^{k_l}{K((\mathbf{x}-\mathbf{x}_i^l)./\mathbf{h}_l)}  \\
& = \arg \max_{l\in Y} \sum_{i=1}^{k_l}{K(\mathbf{x}-\mathbf{x}_i^l)}. \quad assume \quad h_l=\mathbf{1}
\end{aligned}
\label{dwknnrule}
\end{equation}
where $\mathbf{x}_i^l$ is the $i$th sample of class $l$ in the neighborhood. Thus, it can be seen as a DW-kNN by assigning a weight $K(\mathbf{x}-\mathbf{x}_i)$ to sample $\mathbf{x}_i$. A different kernel function $K(\cdot)$ corresponds to a different weight function.

\textbf{Local mean method (LMM).}
An LMM computes a local center of the neighborhood of the query sample for each class, and then minimizes the distance between the local centers and the query sample. Usually, an equal number of nearest neighbors are selected from the corresponding class to estimate the corresponding local center. A number of articles \cite{hotta2004pattern,mitani2006local,li2008nearest,gou2012local} have presented classifiers of this kind.

An LMM is a special case of LPM-BC when using LGA with the same covariance matrix $\Sigma$ for each class. If we select the neighborhood $R_l(\mathbf{x})$ for class $l$ such that there are $k$ samples from class $l$ in $R_l(\mathbf{x})$, i.e. $k_l=k$ is constant for each class, the LPM-BC rule can be transformed as
\begin{equation}
\small
\begin{aligned}
\hat {l} & = \arg \max_{l\in Y} k\cdot g(\mathbf{x};\hat{\mu}_l,\Sigma)  \\
& = \arg \max_{l\in Y} \frac{1}{\sqrt{(2\pi)^d |\Sigma|}} \exp\{-\frac{(\mathbf{x}-\mu_l)^T \Sigma^{-1} (\mathbf{x}-\mu_l)}{2}\}  \\
& = \arg \min_{l\in Y} (\mathbf{x}-\mu_l)^T \Sigma^{-1} (\mathbf{x}-\mu_l)  \\
\end{aligned}
\end{equation}
where $\mu_l$ is the local center estimated from the $k$ samples in the corresponding local region. $(\mathbf{x}-\mu_l)^T \Sigma^{-1} (\mathbf{x}-\mu_l)$ is the Mahalanobis distance between $\mathbf{x}$ and $\mu_l$. If $\Sigma$ is further assigned an identity matrix, the Mahalanobis distance reduces to the Euclidean distance; that is, this rule assigns the query sample to the class whose local center is closest to the query sample.

\textbf{Local distribution based kNN (LD-kNN).}
We presented the LD-kNN method \cite{mao2015nearest} where the local distribution of each class is assumed Gaussian and the mean and covariance matrix is estimated from the samples in the corresponding neighborhood. An LD-kNN rule is essentially an LPM-BC with LGA in an identical neighborhood for all classes.

\section{Experiments}\label{sec:5}

\subsection{Experimental Setting}
To evaluate the performance of LPM-BC, experiments are performed on 16 benchmark datasets from the well-known UCI machine learning repository \cite{Bache+Lichman:2013}. Detailed information of the datasets is summarized in Table \ref{dataset}. For each dataset, we conduct the following setup.
\begin{table}[!t]
  \centering
    \begin{tabular}{lccc}
    \toprule
    Datasets & \#Samples & \#Features & \#Classes \\
    \midrule
    Blood & 748   & 4     & 2 \\
    Bupaliver & 345   & 6     & 2 \\
    Climate & 540   & 18    & 2 \\
    Diabete & 768   & 8     & 2 \\
    Haberman & 306   & 3     & 2 \\
    Heart & 270   & 13    & 2 \\
    Image & 2310  & 19    & 7 \\
    Ionosphere & 351   & 34    & 2 \\
    Iris  & 150   & 4     & 3 \\
    Libras & 360   & 90    & 15 \\
    Parkinson & 195   & 22    & 2 \\
    Seeds & 210   & 7     & 3 \\
    Sonar & 208   & 60    & 2 \\
    spectf & 267   & 44    & 2 \\
    Vertebral & 310   & 6     & 3 \\
    Wine  & 178   & 13    & 3 \\
    \bottomrule
    \end{tabular}%
  \caption{Dataset information}
  \label{dataset}%
\end{table}%

\textbf{Cross test:} each dataset is randomly stratified into 5 folds; for each iteration, 4 folds constitute the training set $Tr$, the remaining one fold is the test set $Te$. The classification performance is assessed on $Te$; for the 5 folds, the performances on the 5 $Te$ is averaged.

\textbf{Normalization:} each feature is normalized over $Tr$ to have mean 0 and standard deviation 1, and $Te$ are processed with the corresponding parameters.

\textbf{Parameter selection:} the neighborhood size and the LPM assumption are two parameters of LPM-BC, for a test sample $\mathbf{x}$, the neighborhood associated with each class $R_l(\mathbf{x})$ is selected so that it has $k_l=k$ samples in the corresponding class and in $Tr$, parameter $k$ can indicate the neighborhood size. The distance metric to construct the neighborhood is selected Chebychev distance, since it can form a hypercubical neighborhood where the local CCIA would be more likely to hold and where the calculation of integration or volume is reduced. The neighborhood size and the LPM assumption is selected via an internal 4-fold cross validation method on $Tr$. In the internal cross validation, neighborhood size $k$ is selected from $\{1,0.1N_m,0.2N_m,\cdots,N_m\}$, where $N_m$ is the minimum number of samples among all classes; and the LPM assumption is selected among LUA, LGA and LCA. In addition, although an LPM can model the dependencies among features, feature-independent LPMs are established in our experiments based on Theorem \ref{independence} to facilitate the computation.

\textbf{Performance evaluation:} the classification performance is evaluated by accuracy (ACC) and mean square error (MSE) \cite{zhong2013accurate}. To avoid bias, the 5-fold cross test is implemented 8 times and the performances are averaged.

\textbf{Competing classifiers:} the following classifiers are also implemented in our experiments to for comparison. G-NBC and K-NBC\footnote{the global NBC respectively using Gaussian model assumption \cite{duda2012pattern} and KDE \cite{john1995estimating}}, NNBC \cite{wang2014non}, TAN \cite{friedman1997bayesian}, NBTree \cite{kohavi1996scaling}, locally weighted naive Bayes (LWNB) \cite{frank2002locally}, V-kNN, local mean method Categorical Average Pattern (CAP) \cite{hotta2004pattern} and LD-kNN \cite{mao2015nearest}.

%

\subsection{Experimental Results}

\begin{table*}[!t]
  \centering
    \begin{tabular}{c|l|cccccccccc}
    \toprule
          & Datasets & LPM-BC & G-NBC & K-NBC & NNBC  & TAN   & NBTree & LWNB  & V-kNN   & CAP   & LD-kNN \\
    \midrule
    \multicolumn{1}{c|}{\multirow{16}[2]{*}{MSE}} & Blood & 0.1591  & 0.1681  & 0.1657  & \textbf{0.1582 } & 0.1610  & 0.1587  & 0.1604  & 0.1591  & 0.1711  & 0.1622  \\
    \multicolumn{1}{c|}{} & Bupaliver & \textbf{0.2160 } & 0.2595  & 0.2198  & 0.2474  & 0.2443  & 0.2368  & 0.2347  & 0.2331  & 0.2291  & 0.2276  \\
    \multicolumn{1}{c|}{} & Climate & 0.0458  & 0.0399  & 0.0570  & 0.0871  & 0.0527  & 0.0506  & \textbf{0.0396 } & 0.0709  & 0.0652  & 0.0458  \\
    \multicolumn{1}{c|}{} & Diabete & 0.1678  & 0.1779  & 0.1971  & 0.1887  & 0.1725  & 0.1799  & 0.1724  & 0.1832  & \textbf{0.1677 } & 0.1814  \\
    \multicolumn{1}{c|}{} & Haberman & \textbf{0.1785 } & 0.1918  & 0.2527  & 0.2000  & 0.1823  & 0.1840  & 0.1853  & \textbf{0.1785 } & 0.1939  & 0.1962  \\
    \multicolumn{1}{c|}{} & Heart & 0.1391  & 0.1334  & 0.1514  & 0.1722  & 0.1345  & 0.1961  & \textbf{0.1320 } & 0.1886  & 0.1389  & 0.1416  \\
    \multicolumn{1}{c|}{} & Image & \textbf{0.0341 } & 0.1841  & 0.1387  & 0.0582  & 0.0385  & 0.0438  & 0.1474  & 0.0412  & 0.0798  & 0.0768  \\
    \multicolumn{1}{c|}{} & Ionosphere & 0.0858  & 0.1401  & 0.0783  & 0.1020  & \textbf{0.0775 } & 0.1067  & 0.0807  & 0.1085  & 0.1175  & 0.0922  \\
    \multicolumn{1}{c|}{} & Iris  & \textbf{0.0308 } & 0.0364  & 0.0338  & 0.0502  & 0.0518  & 0.0549  & 0.0362  & 0.0468  & 0.0648  & 0.0351  \\
    \multicolumn{1}{c|}{} & Libras & 0.1579  & 0.3458  & 0.3167  & 0.1346  & 0.2490  & 0.2783  & 0.2228  & 0.1579  & \textbf{0.1329 } & 0.2229  \\
    \multicolumn{1}{c|}{} & Parkinson & 0.0656  & 0.2898  & 0.2422  & \textbf{0.0422 } & 0.1222  & 0.1126  & 0.2664  & 0.0656  & 0.0616  & 0.1130  \\
    \multicolumn{1}{c|}{} & Seeds & \textbf{0.0432 } & 0.0813  & 0.0735  & 0.0482  & 0.0731  & 0.0803  & 0.0744  & 0.0582  & 0.0584  & 0.0711  \\
    \multicolumn{1}{c|}{} & Sonar & 0.1205  & 0.2869  & 0.2027  & \textbf{0.1093 } & 0.1811  & 0.2188  & 0.2224  & 0.2212  & 0.1541  & 0.1564  \\
    \multicolumn{1}{c|}{} & spectf & 0.1531  & 0.3020  & 0.2460  & 0.1785  & 0.1586  & 0.1893  & 0.2467  & 0.1860  & \textbf{0.1468 } & 0.1589  \\
    \multicolumn{1}{c|}{} & Vertebral & \textbf{0.1135 } & 0.1202  & 0.1317  & 0.1271  & 0.1311  & 0.1478  & 0.1175  & 0.1428  & 0.1410  & 0.1140  \\
    \multicolumn{1}{c|}{} & Wine  & \textbf{0.0069 } & 0.0207  & 0.0201  & 0.0359  & 0.0215  & 0.0491  & 0.0159  & 0.0462  & 0.0200  & 0.0108  \\
    \cline{2-12}
    \multicolumn{1}{c|}{} & Average rank  &  \textbf{2.28} & 7.44 &	6.63 &	5.50 &	5.13 &	6.88 &	5.19 &	5.94 &	5.19 &	4.84 \\
    \midrule
    \multicolumn{1}{c|}{\multirow{16}[2]{*}{ACC}} & Blood & \textbf{0.7925 } & 0.7644  & 0.7575  & 0.7797  & 0.7513  & 0.7767  & 0.7741  & 0.7821  & 0.7884  & 0.7925  \\
    \multicolumn{1}{c|}{} & Bupaliver & \textbf{0.6851 } & 0.5569  & 0.6435  & 0.6322  & 0.5768  & 0.6551  & 0.6261  & 0.6243  & 0.6572  & 0.6783  \\
    \multicolumn{1}{c|}{} & Climate & 0.9414  & 0.9465  & 0.9188  & 0.8940  & 0.9296  & 0.9370  & \textbf{0.9519 } & 0.9162  & 0.9236  & 0.9414  \\
    \multicolumn{1}{c|}{} & Diabete & \textbf{0.7697 } & 0.7564  & 0.7355  & 0.7332  & 0.7448  & 0.7370  & 0.7643  & 0.7318  & 0.7660  & 0.7645  \\
    \multicolumn{1}{c|}{} & Haberman & 0.7464  & 0.7480  & 0.5295  & 0.7308  & 0.7255  & 0.7353  & \textbf{0.7516 } & 0.7452  & 0.7463  & 0.7464  \\
    \multicolumn{1}{c|}{} & Heart & 0.8380  & 0.8366  & 0.7963  & 0.7787  & 0.8000  & 0.7556  & 0.8296  & 0.7458  & \textbf{0.8384 } & 0.8329  \\
    \multicolumn{1}{c|}{} & Image & \textbf{0.9613 } & 0.7972  & 0.8290  & 0.9243  & 0.9515  & 0.9506  & 0.8229  & 0.9484  & 0.9551  & 0.9154  \\
    \multicolumn{1}{c|}{} & Ionosphere & 0.9106  & 0.8429  & 0.9145  & 0.8771  & \textbf{0.9202 } & 0.8775  & 0.9003  & 0.8608  & 0.8764  & 0.9031  \\
    \multicolumn{1}{c|}{} & Iris  & \textbf{0.9625 } & 0.9558  & 0.9617  & 0.9542  & 0.9267  & 0.9200  & 0.9467  & 0.9500  & 0.9575  & 0.9600  \\
    \multicolumn{1}{c|}{} & Libras & 0.8347  & 0.6287  & 0.6568  & 0.8439  & 0.6667  & 0.6806  & 0.7667  & 0.8347  & \textbf{0.8534 } & 0.7707  \\
    \multicolumn{1}{c|}{} & Parkinson & 0.9186  & 0.7000  & 0.7437  & \textbf{0.9444 } & 0.8410  & 0.8821  & 0.7179  & 0.9186  & 0.9135  & 0.8647  \\
    \multicolumn{1}{c|}{} & Seeds & \textbf{0.9488 } & 0.9030  & 0.9065  & 0.9304  & 0.8857  & 0.9048  & 0.9000  & 0.9262  & 0.9274  & 0.9202  \\
    \multicolumn{1}{c|}{} & Sonar & \textbf{0.8756 } & 0.6837  & 0.7656  & 0.8624  & 0.7548  & 0.7500  & 0.7500  & 0.7494  & 0.8210  & 0.8288  \\
    \multicolumn{1}{c|}{} & spectf & 0.8123  & 0.6794  & 0.7351  & 0.7187  & 0.7903  & 0.7566  & 0.7416  & 0.7940  & \textbf{0.8146 } & 0.8100  \\
    \multicolumn{1}{c|}{} & Vertebral & \textbf{0.8464 } & 0.8218  & 0.8085  & 0.8109  & 0.8000  & 0.7968  & 0.8290  & 0.7859  & 0.8081  & 0.8452  \\
    \multicolumn{1}{c|}{} & Wine  & \textbf{0.9916 } & 0.9755  & 0.9768  & 0.9564  & 0.9775  & 0.9438  & 0.9775  & 0.9416  & 0.9790  & 0.9880  \\
    \cline{2-12}
    \multicolumn{1}{c|}{} & Average rank  & \textbf{1.84} &	7.13 &	6.69 &	5.88 &	6.72 &	6.78 &	5.81 &	6.94 &	3.63 &	3.59\\
    \bottomrule
    \end{tabular}%
  \caption{The performance of different classifiers on real datasets in terms of MSE and ACC. The best performance for each dataset is described in bold-face.}
  \label{performance}%
\end{table*}%

\begin{figure}[!t]
\centering
\subfloat[MSE]{\includegraphics[ width=\columnwidth]{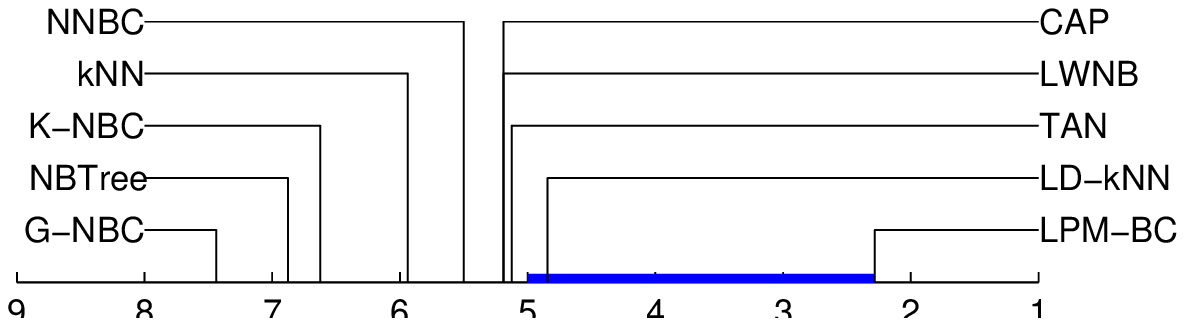}
\label{MSE}}

\subfloat[ACC]{\includegraphics[ width=\columnwidth]{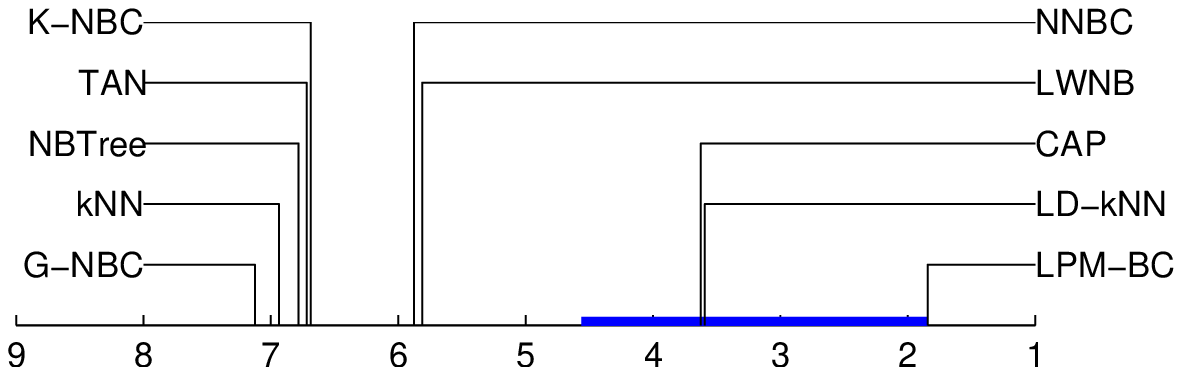}
\label{ACC}}

\caption{Comparisons of LPM-BC against the other classifiers with the
Bonferroni-Dunn test in terms of (a) MSE and (b) ACC. LPM-BC
can significantly ($p < 0.05$) outperform the classifiers with ranks outside the
marked interval.}
\label{multicomp}
\end{figure}

The classification results in terms of ACC and MSE are reported in Table \ref{performance}. We can observe that LPM-BC performs best on 7 and 9 datasets respectively in terms of MSE and ACC, more than all other classifiers. The average rank of LPM-BC on these datasets is respective 2.28 and 1.84, both lower than all other classifiers. These results imply that the LPM-BC can be flexible for various classification problems through tuning the neighborhood size and the corresponding LPM assumption. We employ Friedman tests \cite{hollander1999nonparametric,Dem2006Statistical} for multiple comparisons among these classifiers. The $p$ values in terms of MSE and ACC, respectively, are $2.14\times 10^{-4}$ and $3.86\times 10^{-8}$, both much less than 0.01; it indicates significant difference among these 10 classifiers. We further use the post-hoc Bonferroni-Dunn test \cite{Dem2006Statistical} to reveal the differences among the classifiers. Figure \ref{multicomp} shows the results of the Bonferroni-Dunn test that the other classifiers are compared to LPM-BC. The results indicate that, in terms of MSE and ACC, LPM-BC can significantly ($p<0.05$) outperform all classifiers except LD-kNN and CAP; and that the data is insufficient to discriminate the advantages of LPM-BC over LD-kNN.

\begin{figure}[!t]
\centering
\subfloat{\includegraphics[width=0.9\columnwidth]{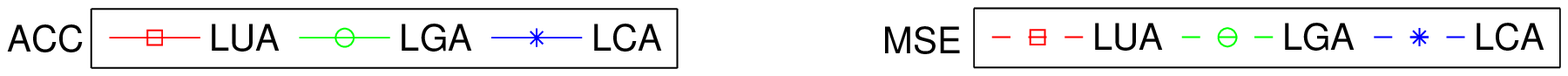}}
\addtocounter{subfigure}{-1}
\vspace{-0.15in}

\subfloat[ionosphere]{\includegraphics[width=0.5\columnwidth]{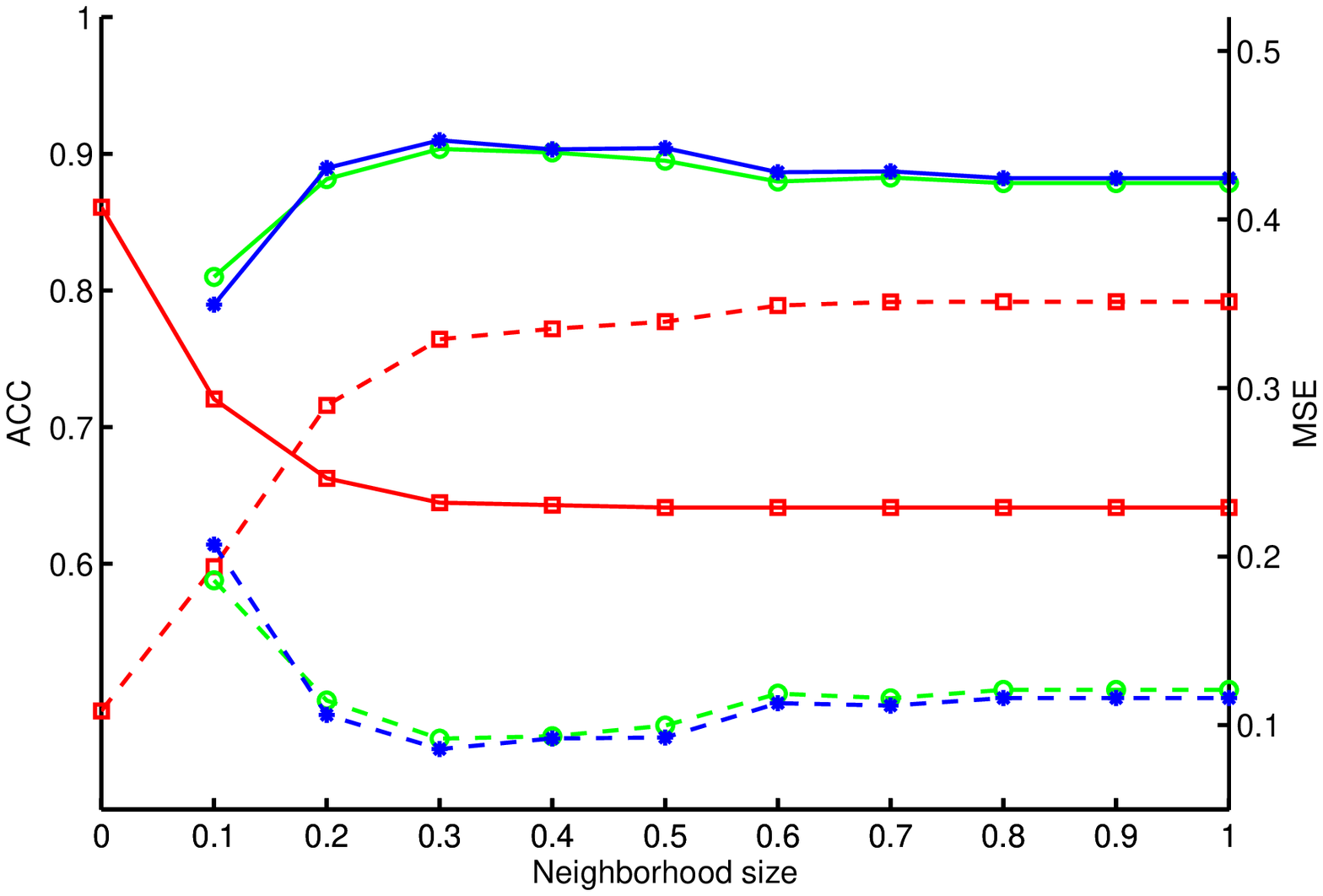}
\label{ionosphere}}
\subfloat[blood]{\includegraphics[width=0.5\columnwidth]{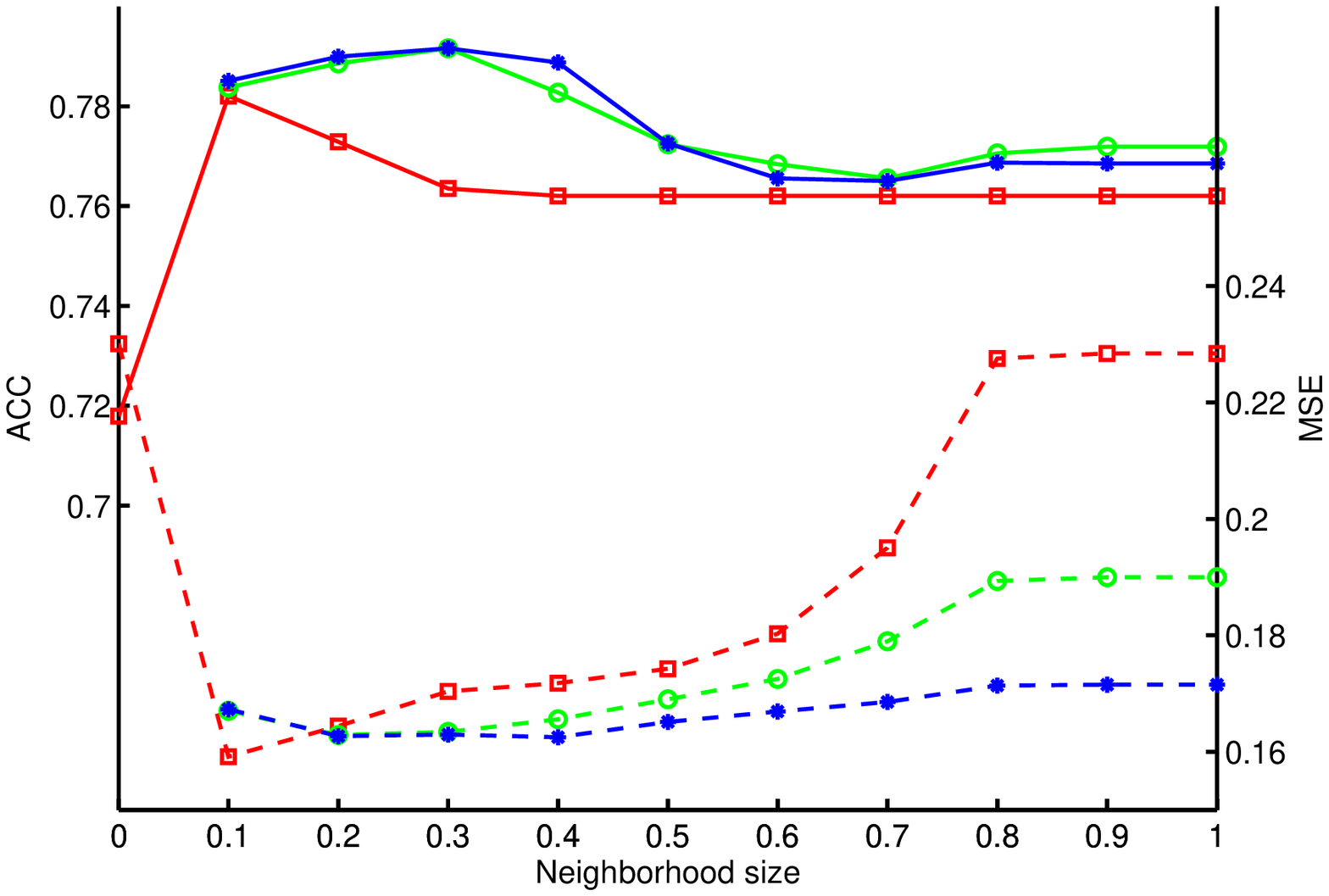}
\label{blood}}

\subfloat[sonar]{\includegraphics[width=0.5\columnwidth]{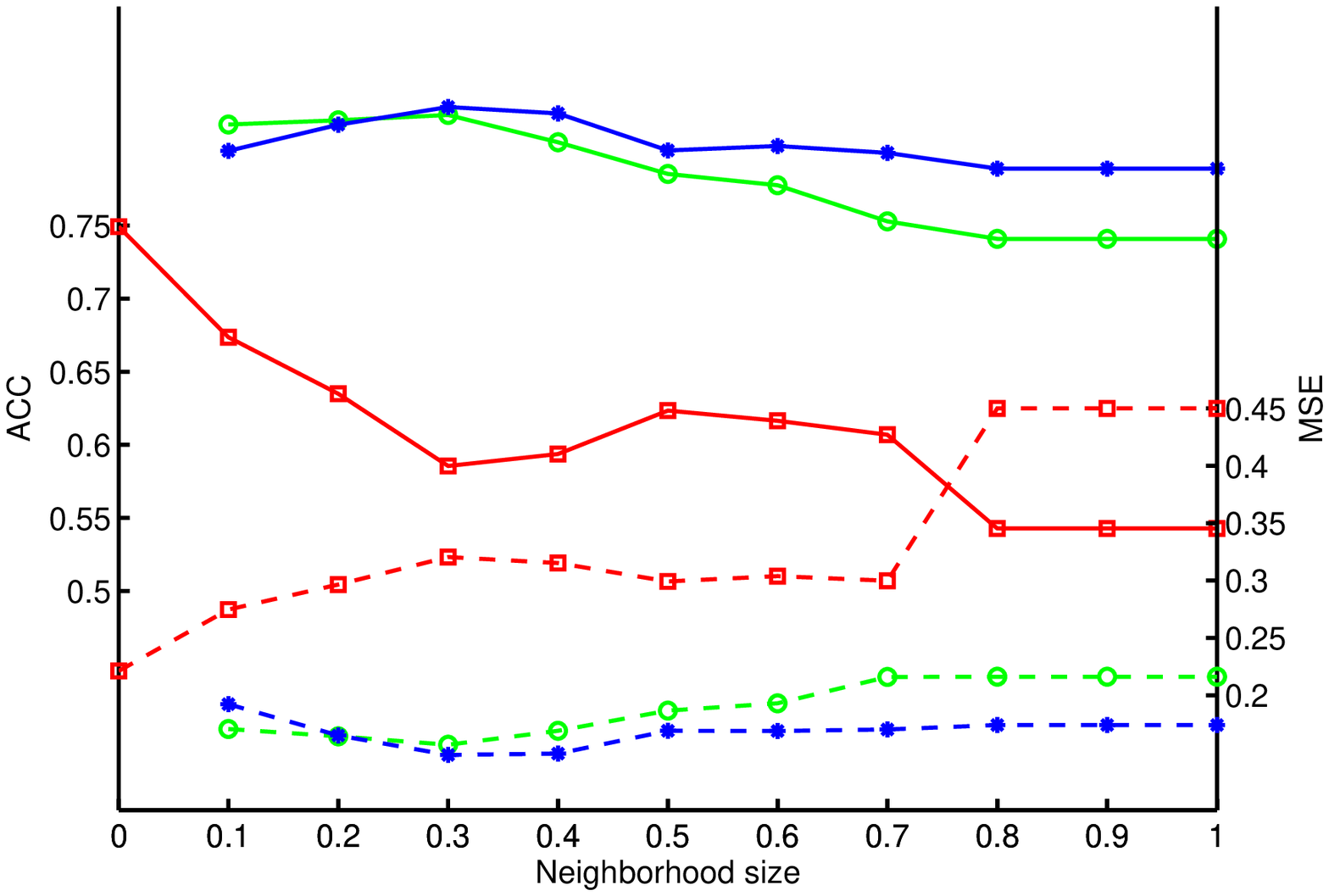}
\label{sonar}}
\subfloat[bupaliver]{\includegraphics[width=0.5\columnwidth]{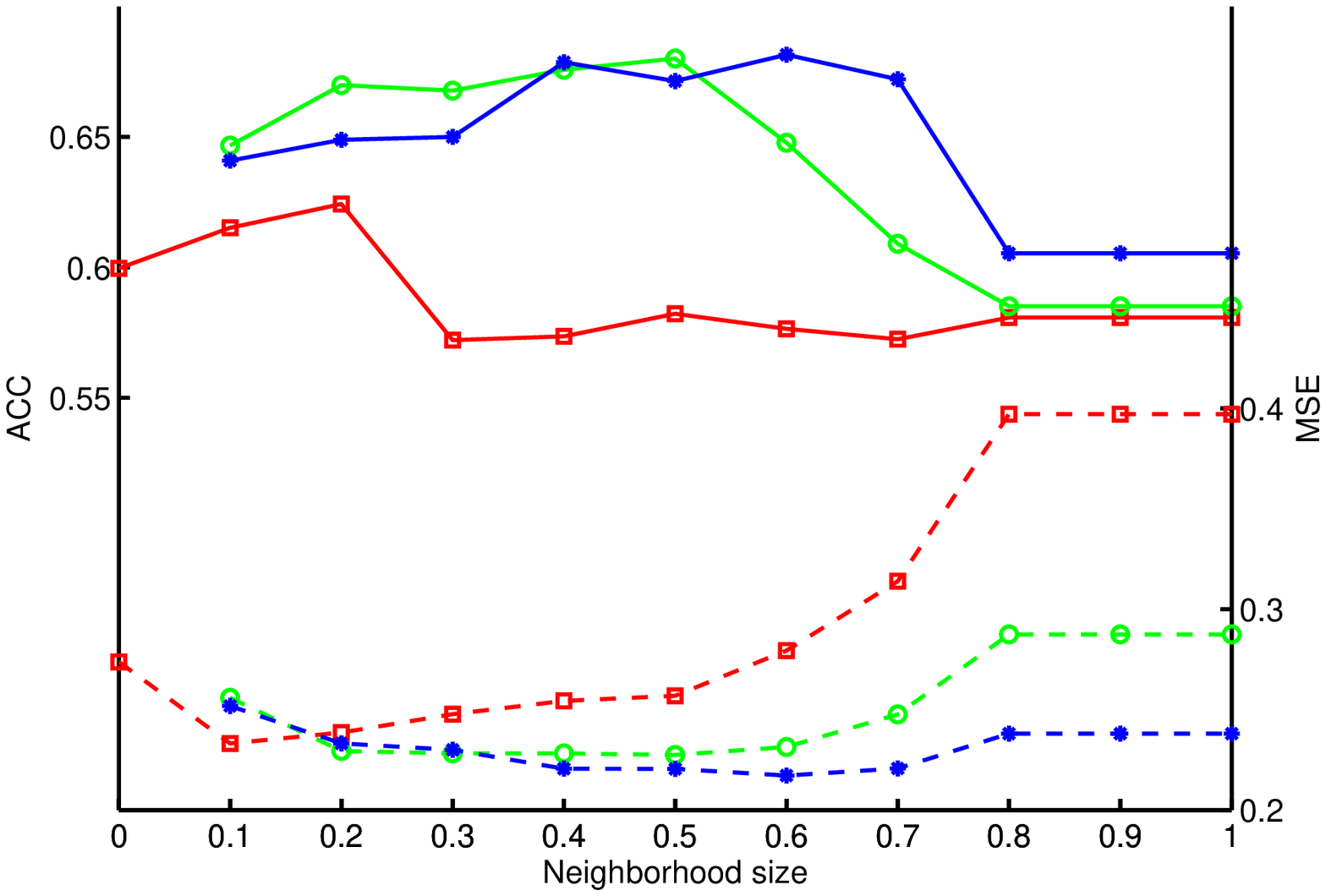}
\label{bupaliver}}

\subfloat[spectf]{\includegraphics[width=0.5\columnwidth]{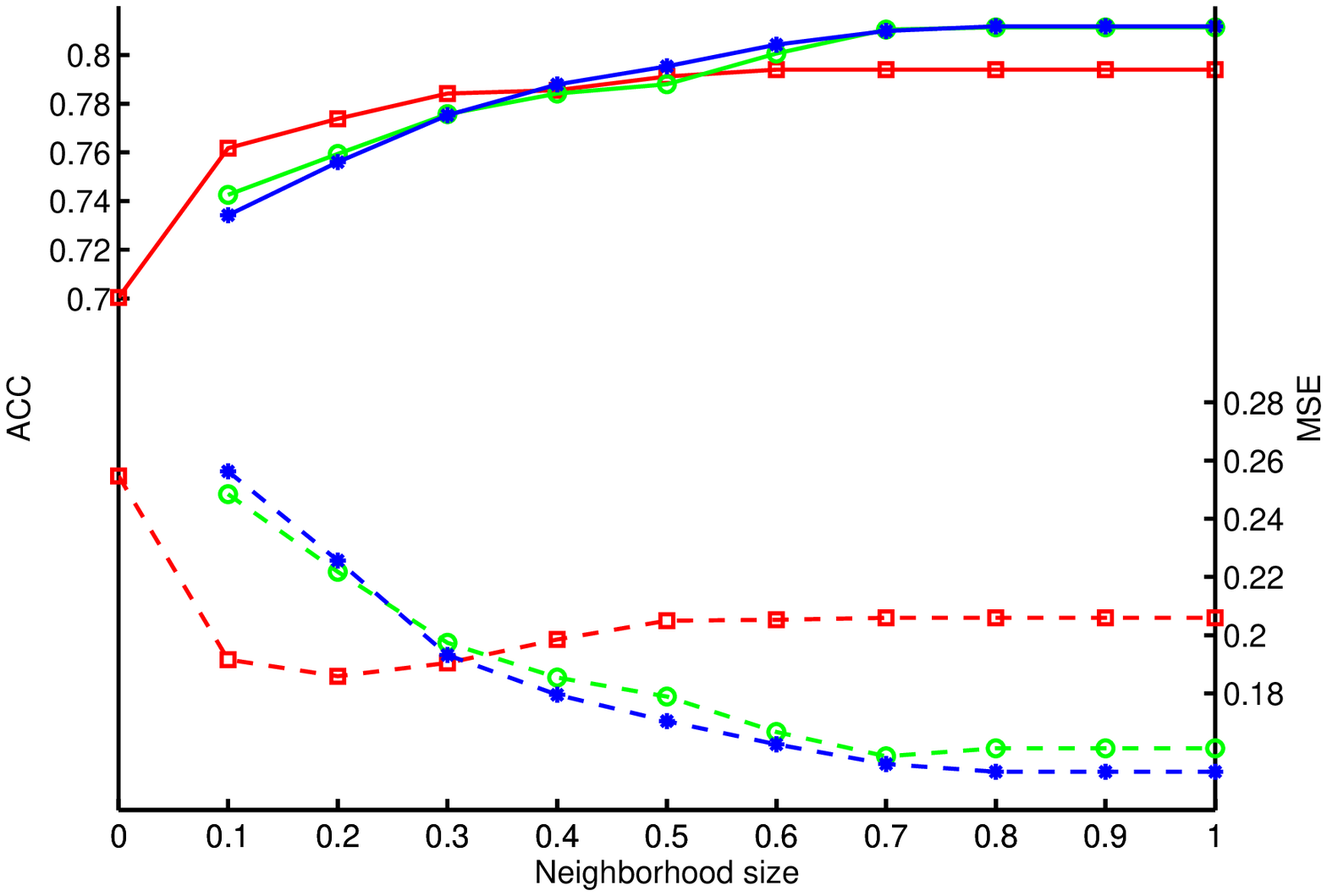}
\label{spectf}}
\subfloat[synthetic($C=0$)]{\includegraphics[width=0.5\columnwidth]{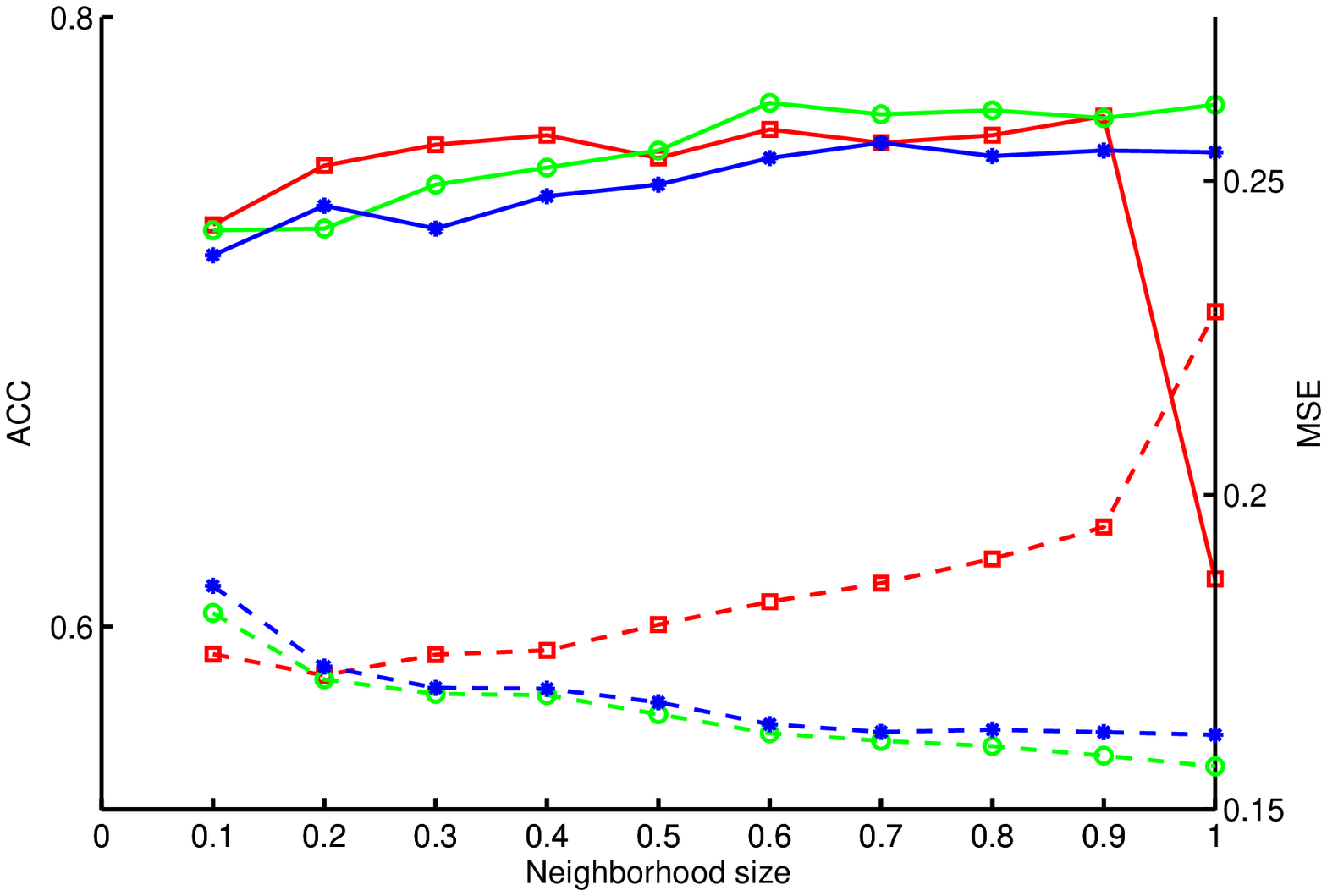}
\label{gaussian0}}

\subfloat[synthetic($C=0.5$)]{\includegraphics[width=0.5\columnwidth]{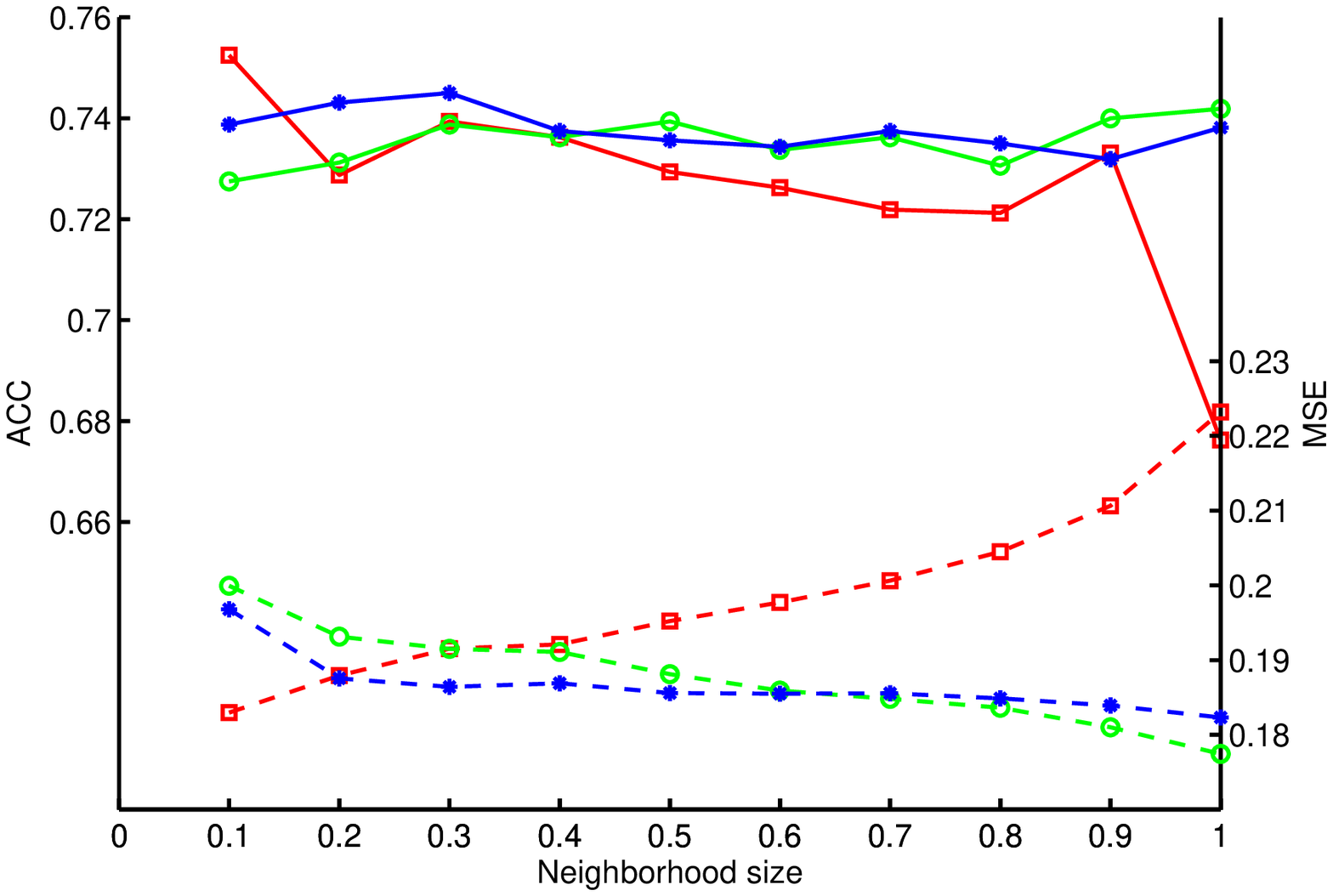}
\label{gaussian0.5}}
\subfloat[synthetic($C=1$)]{\includegraphics[width=0.5\columnwidth]{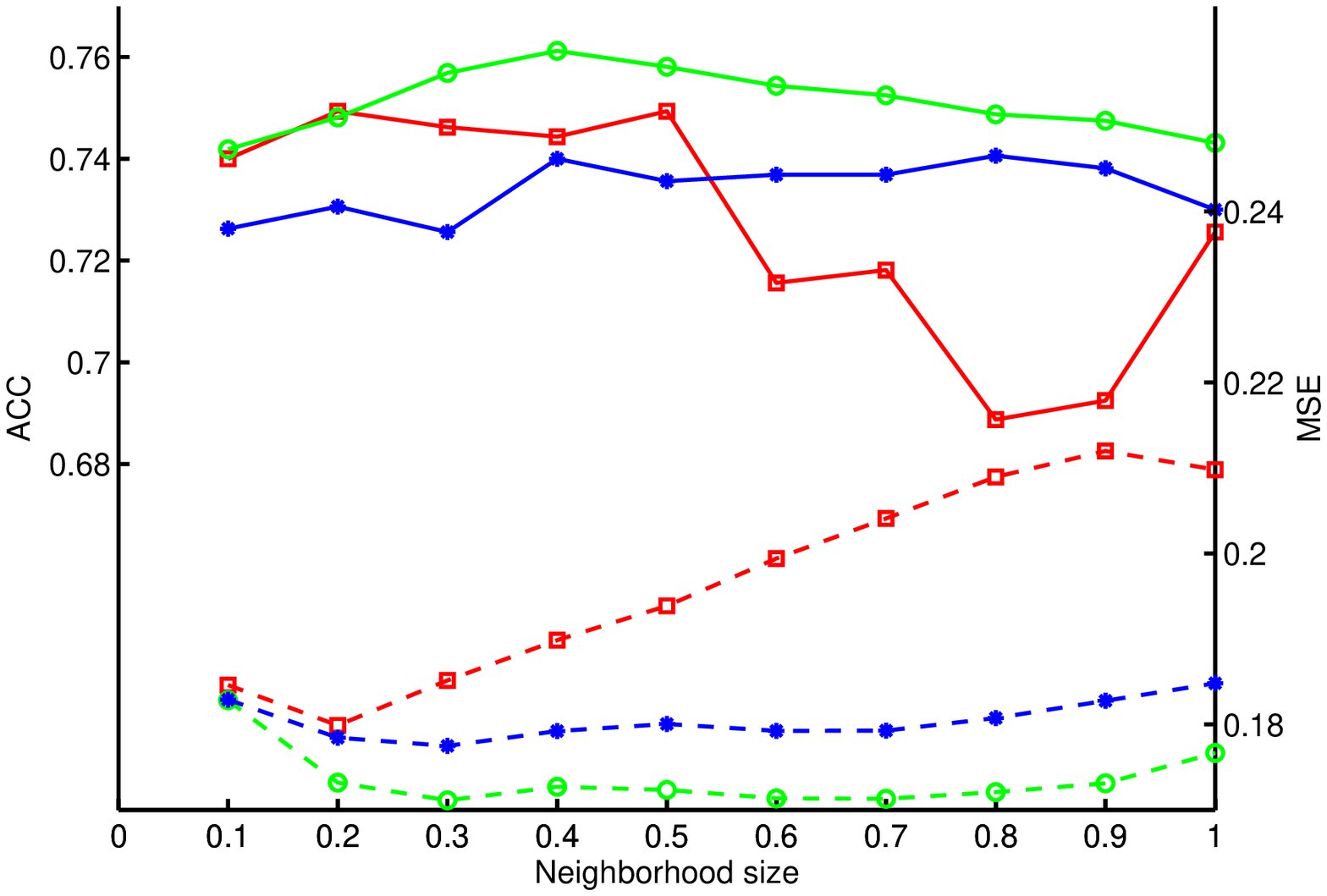}
\label{gaussian1}}

\subfloat[synthetic($C=1.5$)]{\includegraphics[width=0.5\columnwidth]{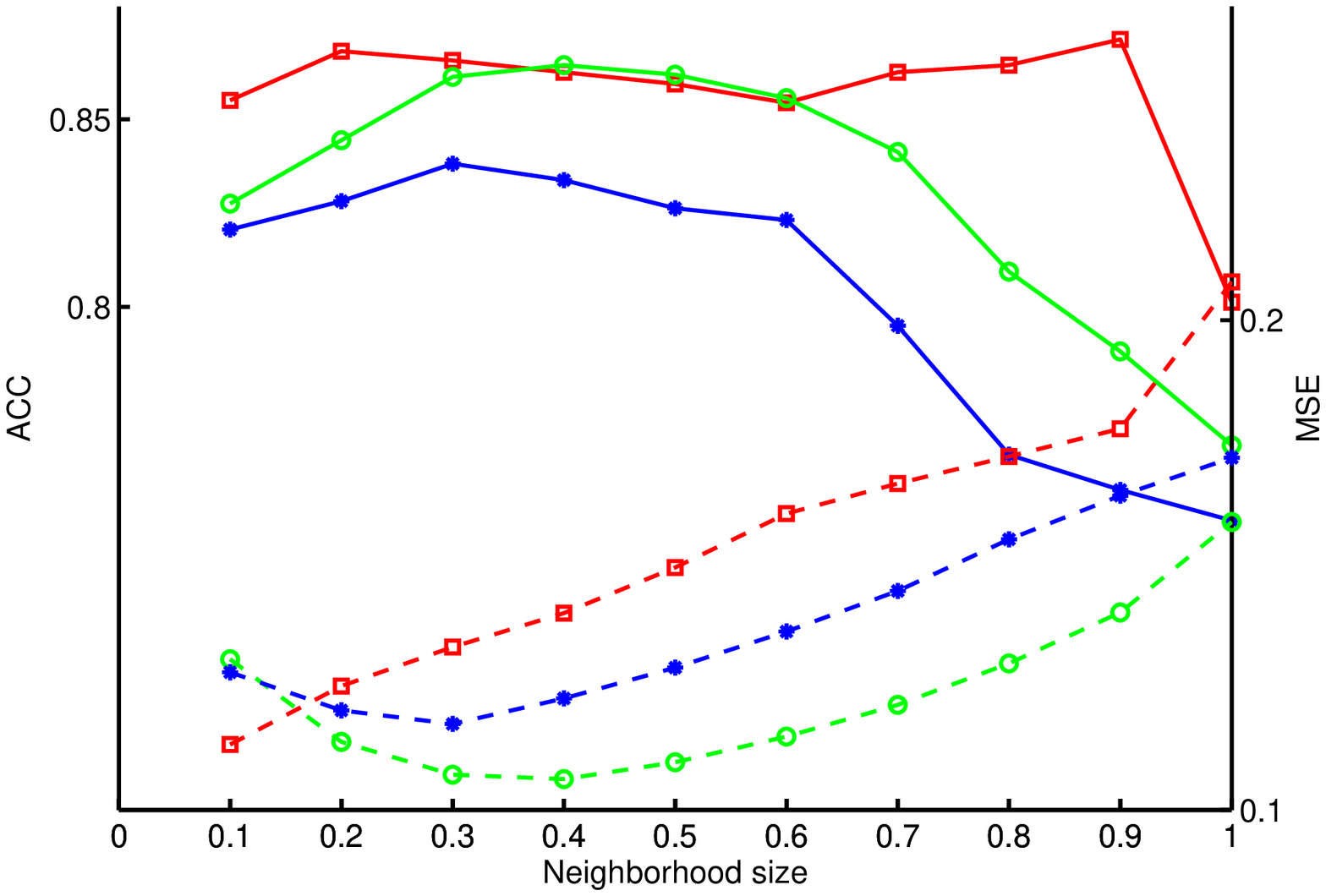}
\label{gaussian1.5}}
\subfloat[synthetic($C=2$)]{\includegraphics[width=0.5\columnwidth]{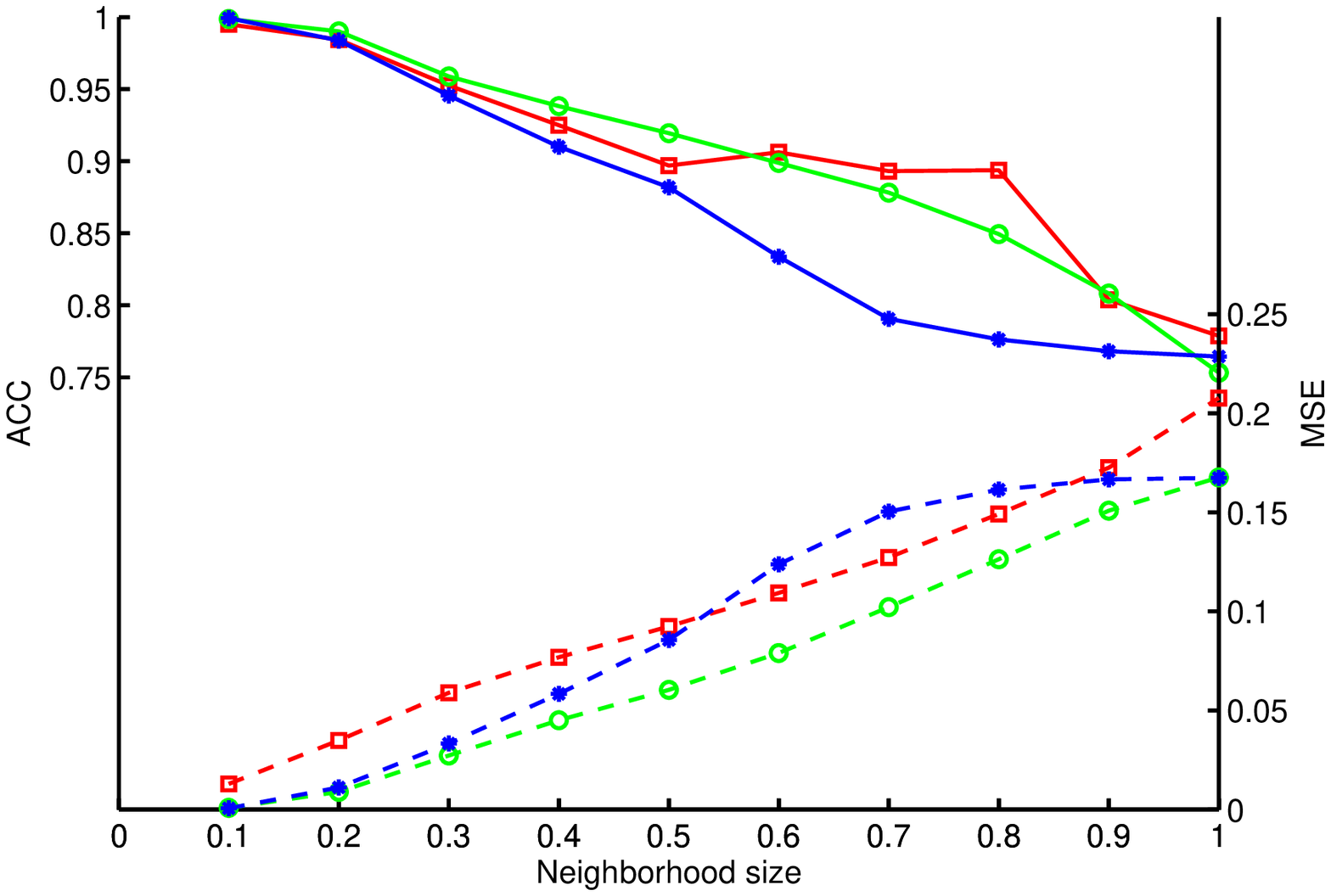}
\label{gaussian2}}

\caption{Classification performances (ACC and MSE) vary with the neighborhood size on several real and synthetic datasets. }
\label{variation}
\end{figure}

\subsection{Parameter Analysis}
The neighborhood size and LPM assumption can influence the classification performance as discussed before. In this experiment, we vary the neighborhood size and take different LPM assumptions to test the performance of LPM-BC on both real and synthetic datasets. Each synthetic dataset is for a 2-dimensional binary classification problem; each of the two classes consists of 100 samples from Gaussian distribution respectively with center $\mu_1=[0,1]$ and $\mu_2=[0,-1]$, and the covariance matrix $\Sigma_1=\Sigma_2=\begin{bmatrix}2 & C \\ C & 2 \end{bmatrix}$; $C$ is selected from $\{0,0.5,1,1.5,2\}$ to construct 5 datasets. The performance curves of ACC and MSE on some representative datasets are shown in Figure \ref{variation}.  From Figure \ref{variation} we can observe a few general trends.
     (1) A simple LPM (e.g. LUA) usually favors a small neighborhood; while a complex LPM (e.g. LCA) usually favors a larger neighborhood.
     (2) LCA does not always perform best in the whole sample space; it depends on the inherent complexity of the true distribution. In real-world problems, the inherent distribution is complex, the LCA is more effective than LGA in the whole sample space; while in the synthetic, the inherent distribution is Gaussian, not so complex, LCA is less effective than LGA.
     (3) LPM-BC usually gets the best performance in a medium-sized neighborhood with a certain LPM for real-world problems.

\textbf{Feature independency.} On the synthetic datasets, the covariance $C$ can describe the dependency between the two features. From Figure \ref{gaussian0} to \ref{gaussian2}, we can see that with the dependency increases, LPM-BC with a larger neighborhood size becomes increasingly ineffective; the superiority of LPM-BC with a small neighborhood to that with a large neighborhood becomes more and more obvious. The results validate Theorem \ref{independence} and indicate that LPM-BC can be a promising Bayesian classifier by relaxing the fundamental CCIA to a local region.

\section{Conclusion}\label{sec:6}
In this paper, we proposed implementing Bayesian classification based on a local probabilistic model. The idea is to transform the estimation of global distribution into the estimation in a local region where the distribution should be simple. LPM-BC is a compromise between parametric and non-parametric method. In the neighborhood of the query sample we assume a parametric probabilistic model while in the whole sample space, it is non-parametric. If the neighborhood is small, it is inclined to be non-parametric and vice versa. Through tuning the neighborhood size, we can control the trade-off between parametric and non-parametric. Also, the LPM-BC can be viewed as a generalized local classification method. Through specifying the local region and LPM, it can be specialized to various local classifiers. Thus, it should be more effective if an appropriate neighborhood and LPM are selected. We have discussed three kinds of LPMs in this
paper, other probabilistic models can also be assumed; the rough rules of neighborhood and LPM selection have been discussed; however, to generate a general selection rule of neighborhood and the corresponding LPM requires further investigation in our future research.

\section*{Acknowledgements}
This work was supported by the National Basic Research Program of China (2014CB744600), the National Natural Science Foundation of China (61402211, 61063028 and 61210010).





\bibliographystyle{IEEEtran}
\bibliography{LD}

\begin{thebibliography}{10}
\providecommand{\url}[1]{#1}
\csname url@samestyle\endcsname
\providecommand{\newblock}{\relax}
\providecommand{\bibinfo}[2]{#2}
\providecommand{\BIBentrySTDinterwordspacing}{\spaceskip=0pt\relax}
\providecommand{\BIBentryALTinterwordstretchfactor}{4}
\providecommand{\BIBentryALTinterwordspacing}{\spaceskip=\fontdimen2\font plus
\BIBentryALTinterwordstretchfactor\fontdimen3\font minus
  \fontdimen4\font\relax}
\providecommand{\BIBforeignlanguage}[2]{{%
\expandafter\ifx\csname l@#1\endcsname\relax
\typeout{** WARNING: IEEEtran.bst: No hyphenation pattern has been}%
\typeout{** loaded for the language `#1'. Using the pattern for}%
\typeout{** the default language instead.}%
\else
\language=\csname l@#1\endcsname
\fi
#2}}
\providecommand{\BIBdecl}{\relax}
\BIBdecl

\bibitem{duda2012pattern}
R.~O. Duda, P.~E. Hart, and D.~G. Stork, \emph{Pattern classification}.\hskip
  1em plus 0.5em minus 0.4em\relax John Wiley \& Sons, 2012.

\bibitem{domingos1996beyond}
P.~Domingos and M.~Pazzani, ``Beyond independence: Conditions for the
  optimality of the simple bayesian classifier,'' in \emph{Proc. 13th Intl.
  Conf. Machine Learning}, 1996, pp. 105--112.

\bibitem{hand2001idiot}
D.~J. Hand and K.~Yu, ``Idiot's bayes-not so stupid after all?''
  \emph{International statistical review}, vol.~69, no.~3, pp. 385--398, 2001.

\bibitem{friedman1997bayesian}
N.~Friedman, D.~Geiger, and M.~Goldszmidt, ``Bayesian network classifiers,''
  \emph{Machine learning}, vol.~29, no. 2-3, pp. 131--163, 1997.

\bibitem{cooper1990computational}
G.~F. Cooper, ``The computational complexity of probabilistic inference using
  bayesian belief networks,'' \emph{Artificial intelligence}, vol.~42, no.~2,
  pp. 393--405, 1990.

\bibitem{dagum1993approximating}
P.~Dagum and M.~Luby, ``Approximating probabilistic inference in bayesian
  belief networks is np-hard,'' \emph{Artificial intelligence}, vol.~60, no.~1,
  pp. 141--153, 1993.

\bibitem{webb2005not}
G.~I. Webb, J.~R. Boughton, and Z.~Wang, ``Not so naive bayes: aggregating
  one-dependence estimators,'' \emph{Machine learning}, vol.~58, no.~1, pp.
  5--24, 2005.

\bibitem{wang2014non}
X.-Z. Wang, Y.-L. He, and D.~D. Wang, ``Non-naive bayesian classifiers for
  classification problems with continuous attributes,'' \emph{Cybernetics, IEEE
  Transactions on}, vol.~44, no.~1, pp. 21--39, 2014.

\bibitem{bottou1992local}
L.~Bottou and V.~Vapnik, ``Local learning algorithms,'' \emph{Neural
  computation}, vol.~4, no.~6, pp. 888--900, 1992.

\bibitem{huang2005local}
K.~Huang, H.~Yang, I.~King, and M.~R. Lyu, ``Local learning vs. global
  learning: An introduction to maxi-min margin machine,'' in \emph{Support
  vector machines: theory and applications}.\hskip 1em plus 0.5em minus
  0.4em\relax Springer, 2005, pp. 113--131.

\bibitem{wu2006local}
M.~Wu and B.~Sch{\"o}lkopf, ``A local learning approach for clustering,'' in
  \emph{Advances in neural information processing systems}, 2006, pp.
  1529--1536.

\bibitem{bishop2006pattern}
C.~M. Bishop \emph{et~al.}, \emph{Pattern recognition and machine
  learning}.\hskip 1em plus 0.5em minus 0.4em\relax springer New York, 2006,
  vol.~1.

\bibitem{hjort1996locally}
N.~L. Hjort and M.~Jones, ``Locally parametric nonparametric density
  estimation,'' \emph{The Annals of Statistics}, pp. 1619--1647, 1996.

\bibitem{loader1996local}
C.~R. Loader \emph{et~al.}, ``Local likelihood density estimation,'' \emph{The
  Annals of Statistics}, vol.~24, no.~4, pp. 1602--1618, 1996.

\bibitem{vincent2003locally}
P.~Vincent, Y.~Bengio \emph{et~al.}, ``Locally weighted full covariance
  gaussian density estimation,'' Technical report 1240, Tech. Rep., 2003.

\bibitem{cover1967nearest}
T.~Cover and P.~Hart, ``Nearest neighbor pattern classification,''
  \emph{Information Theory, IEEE Transactions on}, vol.~13, no.~1, pp. 21--27,
  1967.

\bibitem{larose2006k}
D.~T. Larose and C.~D. Larose, ``k-nearest neighbor algorithm,''
  \emph{Discovering Knowledge in Data: An Introduction to Data Mining, Second
  Edition}, pp. 149--164, 2006.

\bibitem{zhang2006svm}
H.~Zhang, A.~Berg, M.~Maire, and J.~Malik, ``Svm-knn: Discriminative nearest
  neighbor classification for visual category recognition,'' in \emph{Computer
  Vision and Pattern Recognition, 2006 IEEE Computer Society Conference on},
  vol.~2.\hskip 1em plus 0.5em minus 0.4em\relax IEEE, 2006, pp. 2126--2136.

\bibitem{cheng2007localized}
H.~Cheng, P.-N. Tan, and R.~Jin, ``Localized support vector machine and its
  efficient algorithm.'' in \emph{SDM}.\hskip 1em plus 0.5em minus 0.4em\relax
  SIAM, 2007, pp. 461--466.

\bibitem{blanzieri2008nearest}
E.~Blanzieri and F.~Melgani, ``Nearest neighbor classification of remote
  sensing images with the maximal margin principle,'' \emph{Geoscience and
  Remote Sensing, IEEE Transactions on}, vol.~46, no.~6, pp. 1804--1811, 2008.

\bibitem{ladicky2011locally}
L.~Ladicky and P.~Torr, ``Locally linear support vector machines,'' in
  \emph{Proceedings of the 28th International Conference on Machine Learning
  (ICML-11)}, 2011, pp. 985--992.

\bibitem{kohavi1996scaling}
R.~Kohavi, ``Scaling up the accuracy of naive-bayes classifiers: A
  decision-tree hybrid.'' in \emph{KDD}.\hskip 1em plus 0.5em minus 0.4em\relax
  Citeseer, 1996, pp. 202--207.

\bibitem{zheng2000lazy}
Z.~Zheng and G.~I. Webb, ``Lazy learning of bayesian rules,'' \emph{Machine
  Learning}, vol.~41, no.~1, pp. 53--84, 2000.

\bibitem{frank2002locally}
E.~Frank, M.~Hall, and B.~Pfahringer, ``Locally weighted naive bayes,'' in
  \emph{Proceedings of the Nineteenth conference on Uncertainty in Artificial
  Intelligence}.\hskip 1em plus 0.5em minus 0.4em\relax Morgan Kaufmann
  Publishers Inc., 2002, pp. 249--256.

\bibitem{xie2002snnb}
Z.~Xie, W.~Hsu, Z.~Liu, and M.~L. Lee, ``Snnb: A selective neighborhood based
  naive bayes for lazy learning,'' in \emph{Advances in knowledge discovery and
  data mining}.\hskip 1em plus 0.5em minus 0.4em\relax Springer, 2002, pp.
  104--114.

\bibitem{hu2015bayesian}
B.~Hu, C.~Mao, X.~Zhang, and Y.~Dai, ``Bayesian classification with local
  probabilistic model assumption in aiding medical diagnosis,'' in
  \emph{Bioinformatics and Biomedicine (BIBM), 2015 IEEE International
  Conference on}.\hskip 1em plus 0.5em minus 0.4em\relax IEEE, 2015, pp.
  691--694.

\bibitem{trevor2009elements}
T.~J. Hastie, R.~J. Tibshirani, and J.~H. Friedman, \emph{The elements of
  statistical learning: data mining, inference, and prediction}.\hskip 1em plus
  0.5em minus 0.4em\relax Springer, 2009.

\bibitem{hechenbichler2004weighted}
K.~Hechenbichler and K.~Schliep, ``Weighted k-nearest-neighbor techniques and
  ordinal classification,'' 2004.

\bibitem{hotta2004pattern}
S.~Hotta, S.~Kiyasu, and S.~Miyahara, ``Pattern recognition using average
  patterns of categorical k-nearest neighbors,'' in \emph{Pattern Recognition,
  2004. ICPR 2004. Proceedings of the 17th International Conference on},
  vol.~4.\hskip 1em plus 0.5em minus 0.4em\relax IEEE, 2004, pp. 412--415.

\bibitem{mitani2006local}
Y.~Mitani and Y.~Hamamoto, ``A local mean-based nonparametric classifier,''
  \emph{Pattern Recognition Letters}, vol.~27, no.~10, pp. 1151--1159, 2006.

\bibitem{li2008nearest}
B.~Li, Y.~Chen, and Y.~Chen, ``The nearest neighbor algorithm of local
  probability centers,'' \emph{Systems, Man, and Cybernetics, Part B:
  Cybernetics, IEEE Transactions on}, vol.~38, no.~1, pp. 141--154, 2008.

\bibitem{gou2012local}
J.~Gou, Z.~Yi, L.~Du, and T.~Xiong, ``A local mean-based k-nearest centroid
  neighbor classifier,'' \emph{The Computer Journal}, vol.~55, no.~9, pp.
  1058--1071, 2012.

\bibitem{mao2015nearest}
C.~Mao, B.~Hu, P.~Moore, Y.~Su, and M.~Wang, ``Nearest neighbor method based on
  local distribution for classification,'' in \emph{Advances in Knowledge
  Discovery and Data Mining}.\hskip 1em plus 0.5em minus 0.4em\relax Springer,
  2015, pp. 239--250.

\bibitem{Bache+Lichman:2013}
\BIBentryALTinterwordspacing
K.~Bache and M.~Lichman, ``{UCI} machine learning repository,'' 2013. [Online].
  Available: \url{http://archive.ics.uci.edu/ml}
\BIBentrySTDinterwordspacing

\bibitem{zhong2013accurate}
L.~W. Zhong and J.~T. Kwok, ``Accurate probability calibration for multiple
  classifiers,'' in \emph{Proceedings of the Twenty-Third international joint
  conference on Artificial Intelligence}.\hskip 1em plus 0.5em minus
  0.4em\relax AAAI Press, 2013, pp. 1939--1945.

\bibitem{john1995estimating}
G.~H. John and P.~Langley, ``Estimating continuous distributions in bayesian
  classifiers,'' in \emph{Proceedings of the Eleventh conference on Uncertainty
  in artificial intelligence}.\hskip 1em plus 0.5em minus 0.4em\relax Morgan
  Kaufmann Publishers Inc., 1995, pp. 338--345.

\bibitem{hollander1999nonparametric}
M.~Hollander and D.~A. Wolfe, ``Nonparametric statistical methods,'' \emph{NY
  John Wiley \& Sons}, 1999.

\bibitem{Dem2006Statistical}
Demsar and J.~Ar, ``Statistical comparisons of classifiers over multiple data
  sets,'' \emph{Journal of Machine Learning Research}, vol.~7, no.~1, pp.
  1--30, 2006.

\end{thebibliography}

\appendix
\section{Proof of Theorem 1}
We only prove the cases with two variables. The mathematical expressions of Theorem 1 are as follows.

$A,B$ are two random variables, $\Omega$ is a subregion of the sample space of $(A,B)$. \\
 if $\forall x,y, P(A=x,B=y \vert \Omega)= P(A=x \vert \Omega)P(B=y \vert \Omega)$  \\
 then $\forall\omega  \quad  s.t. \quad \omega \subseteq \Omega,  \omega=\{(A,B):a \leq A \leq m, b \leq B \leq n\}$
         we have $\forall x,y, P(A=x,B=y \vert \omega)= P(A=x \vert \omega)P(B=y \vert \omega)$.

\begin{proof}
because $\omega \subseteq \Omega$, then \\
\begin{equation}
\begin{aligned}
 P(\omega) & = P(\omega|\Omega)P(\Omega) \\
 & = P(a \leq A \leq m, b \leq B \leq n | \Omega)P(\Omega) \\
 & = P(a \leq A \leq m|\Omega) P(b \leq B \leq n | \Omega)P(\Omega)
\end{aligned}
\end{equation}
then, \\
\begin{equation}
\begin{aligned}
P(A=x,B=y \vert \omega) & = \frac{P(A=x,B=y , \omega)}{P(\omega)} \\
& = \frac{P(A=x,B=y , \omega |\Omega)P(\Omega)}{P(\omega)} \\
\end{aligned}
\end{equation}
if $(x,y)\notin \omega$,  $P(A=x,B=y \vert \omega)=0$  \\
if $(x,y)\in \omega$ $P(A=x,B=y , \omega |\Omega)=P(A=x,B=y|\Omega)$\\
\begin{equation}
\begin{aligned}
  P(A=x,B=y \vert \omega) &= \frac{P(A=x|\Omega)P(B=y|\Omega)P(\Omega)}{P(\omega)} \\
  & = \frac{P(A=x|\Omega)P(B=y|\Omega)P(\Omega)}{P(a \leq A \leq m|\Omega) P(b \leq B \leq n | \Omega)}
\end{aligned}
\end{equation}

if $(x,y)\notin \omega$, $P(A=x \vert \omega)P(B=y \vert \omega) =0$. \\
if $(x,y)\in \omega$ \\
\begin{equation}
\begin{aligned}
& P(A=x \vert \omega)P(B=y \vert \omega) \\
& = \frac{P(A=x,\omega)}{P(\omega)} \cdot\frac{P(B=y,\omega)}{P(\omega)} \\
& = \frac{P(A=x,\omega|\Omega)P(\Omega)}{P(\omega)}\cdot \frac{P(B=y,\omega|\Omega)P(\Omega)}{P(\omega)}   \\
& = \frac{P(A=x|\Omega)P(B=y|\Omega)P(\Omega)}{P(a \leq A \leq m|\Omega) P(b \leq B \leq n | \Omega)}
\end{aligned}
\end{equation}

$P(A=x,B=y \vert \omega)= P(A=x \vert \omega)P(B=y \vert \omega)$ holds.

\end{proof}


%
%
%
\end{document}